\documentclass[preprint,12pt]{elsarticle}
\PassOptionsToPackage{linesnumbered, ruled, vlined}{algorithm2e}
\usepackage{array}
\usepackage{booktabs}
\usepackage{makecell}
\usepackage{multirow}
\usepackage{algorithm2e}
\usepackage{amsmath}
\usepackage{amssymb}
\usepackage{graphicx}
\usepackage{subfigure}
\graphicspath{{figures/}}
\usepackage{enumerate}
\usepackage{textcomp}
\usepackage{xpatch}
\usepackage{color}

\providecommand{\tabularnewline}{\\}

\usepackage{url}

\usepackage{hyperref}
\makeatletter
\def\UrlAlphabet{%
      \do\a\do\b\do\c\do\d\do\e\do\f\do\g\do\h\do\i\do\j%
      \do\k\do\l\do\m\do\n\do\o\do\p\do\q\do\r\do\s\do\t%
      \do\u\do\v\do\w\do\x\do\y\do\z\do\A\do\B\do\C\do\D%
      \do\E\do\F\do\G\do\H\do\I\do\J\do\K\do\L\do\M\do\N%
      \do\O\do\P\do\Q\do\R\do\S\do\T\do\U\do\V\do\W\do\X%
      \do\Y\do\Z}
\def\UrlDigits{\do\1\do\2\do\3\do\4\do\5\do\6\do\7\do\8\do\9\do\0}
\g@addto@macro{\UrlBreaks}{\UrlOrds}
\g@addto@macro{\UrlBreaks}{\UrlAlphabet}
\g@addto@macro{\UrlBreaks}{\UrlDigits}

\def\@linkbordercolor{0.25 0.88 0.8}%
\makeatother

\journal{Applied Soft Computing}

\begin{document}

\begin{frontmatter}{}

\title{A Self-adaptive Neuroevolution Approach to Constructing Deep Neural Network Architectures Across Different Types}

\author[Hongbo2]{Zhenhao Shuai}
\ead{dr.shuai@dlmu.edu.cn}

\author[Hongbo2]{Hongbo~Liu\corref{cor1}}
\ead{lhb@dlmu.edu.cn}

\author[Hongbo2]{Zhaolin Wan\corref{cor1}}
\ead{zlwan@dlmu.edu.cn}

\author[ywj]{Wei-Jie Yu}
\ead{yuweijie6@mail.sysu.edu.cn}

\author[ZhangJun1,ZhangJun2,ZhangJun3]{Jun Zhang}
\ead{junzhang@ieee.org}

\cortext[cor1]{Corresponding author.
}

\address[Hongbo2]{College of Artificial Intelligence, Dalian Maritime University, Dalian 116026, China}
\address[ywj]{School of Information Management, Sun Yat-sen University, Guangzhou 510006, China}
\address[ZhangJun1]{Hanyang University, Ansan 15588, Korea}
\address[ZhangJun2]{Victoria University, Melbourne, VIC 8001, Australia}
\address[ZhangJun3]{Chaoyang University of Technology, Taichung 41349, Taiwan}

\begin{abstract}
Neuroevolution has greatly promoted Deep Neural Network (DNN) architecture design and its applications, while there is a lack of methods available across different DNN types concerning both their scale and performance. In this study, we propose a self-adaptive neuroevolution (SANE) approach to automatically construct various lightweight DNN architectures for different tasks. One of the key settings in SANE is the search space defined by cells and organs self-adapted to different DNN types. Based on this search space, a constructive evolution strategy with uniform evolution settings and operations is designed to grow DNN architectures gradually. SANE is able to self-adaptively adjust evolution exploration and exploitation to improve search efficiency. Moreover, a speciation scheme is developed to protect evolution from early convergence by restricting selection competition within species. To evaluate SANE, we carry out neuroevolution experiments to generate different DNN architectures including convolutional neural network, generative adversarial network and long short-term memory. The results illustrate that the obtained DNN architectures could have smaller scale with similar performance compared to existing DNN architectures. Our proposed SANE provides an efficient approach to self-adaptively search DNN architectures across different types. Our Code is available at \url{https://github.com/ACoTAI/ASOC-SANE}.
\end{abstract}

\begin{graphicalabstract} 
\includegraphics[width=1.0\textwidth]{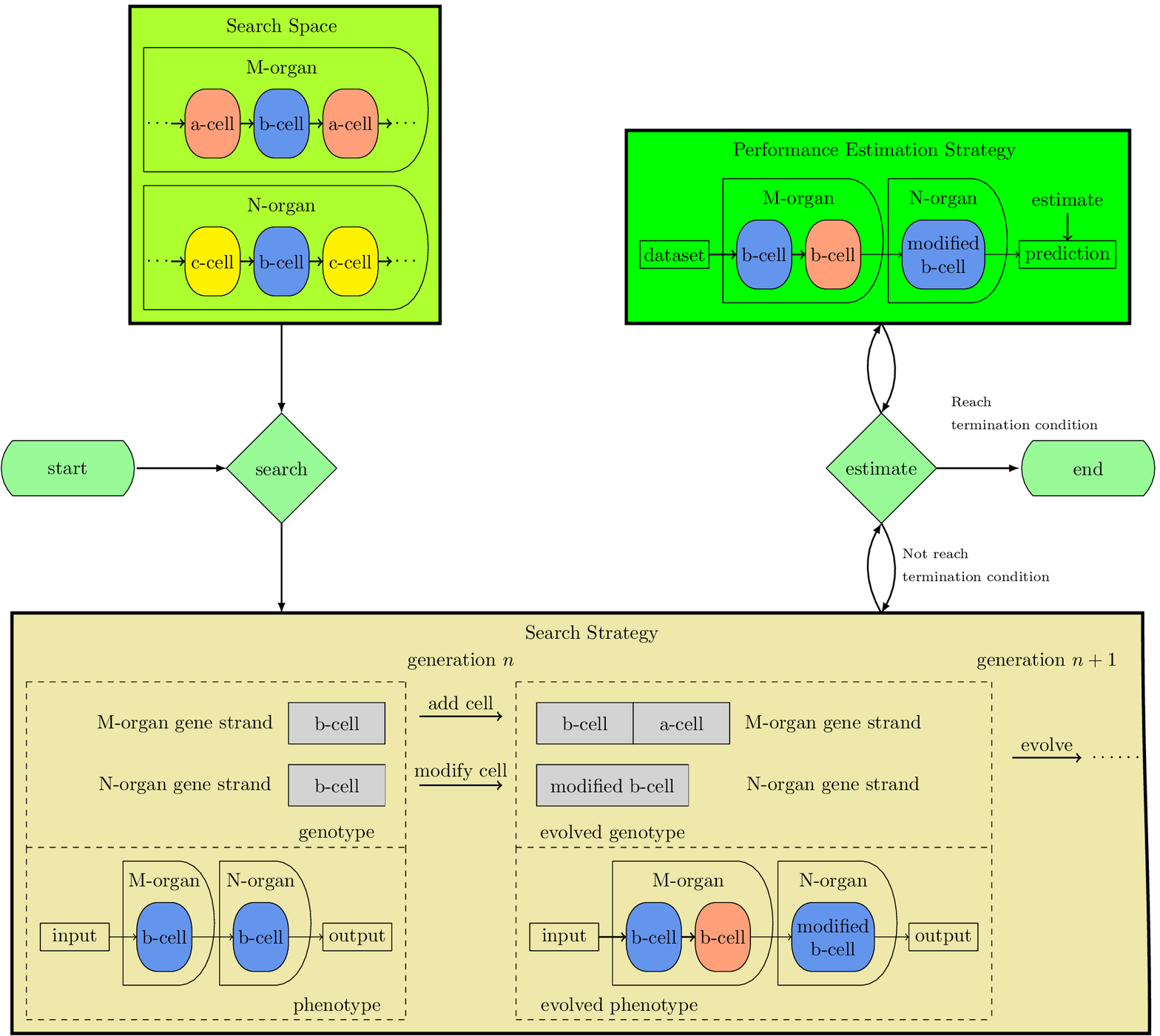}
\end{graphicalabstract}


\end{frontmatter}{}

\section{Introduction}

Deep Neural Network (DNN) has dramatically improved the state-of-the-art performance in many applications, especially in Computer Vision (CV) \cite{lecun_deep_2015}. Most effective DNN architectures are designed manually by human experts, which is time-consuming and requires a high level of expertise. Due to the drastic growth in the demand for DNN architecture design, a research field concerning automating DNN architecture engineering named neural architecture search (NAS) has generated considerable interest in recent years.

Although great progress has been made in NAS research \cite{life_mo_2021, rakhshani_protein_2021, zhang_asnas_2021}, there are still two main issues. (1) The enormous computing consumption has always been a big obstacle. Pioneering NAS studies \cite{zoph_neural_2017, real_large-scale_2017} consume hundreds of graphic processing unit (GPU) days to generate DNN architectures. Therefore, many subsequent studies focus on computing reduction, among which weight sharing \cite{pham_efficient_2018} and one-shot search \cite{guo_oneshot_2020} are two representative strategies. These methods are only helpful to computation relief through improving DNN estimation strategy, however, the key of computing reduction in NAS is to dynamically control its search setting such as the quantity of DNN architectures according to search states. (2) The search automation is restrained within single DNN type. Although researchers have taken much effort to improve prediction performance for specific DNN types and made many great achievements \cite{li_cnn_2021, jiang_optim_2020, wang_adaptive_2021}, these task-oriented methods are not easy to seamlessly transfer across different DNN types. Furthermore, the pursuit of performance without considering the scale of the obtained DNN architectures is not beneficial to the balance between its performance and scale in practice. 

In this paper, we propose a self-adaptive neuroevolution (SANE) approach to automatically construct DNN architectures of different types with near-minimal scale for particular performance standards. According to the similarities among different DNN architectures in their components and connections, we come up with a type-free search space which is capable to self-adapt to different DNN types. Together with type-free search space, we develop a constructive Evolutionary Algorithm (EA), in which DNN architectures (individuals) are the same with minimal necessary components in initial population. As evolution goes on, individuals grow by adding and modifying cells. During evolution, a self-adaptive adjustment mechanism is designed to automatically regulate search exploitation and exploration settings according to the changing search state.

Our main contributions are listed as follows:
\begin{enumerate}[1)]
\item We come up with a type-free search space and uniform evolution search strategy to construct different types of DNN architectures. In SANE, DNN architectures are considered as entities composed by cells and organs in a type-free search space, so that a set of uniform evolution operations is able to construct various DNN architectures.
\item We design a self-adaptive adjustment mechanism to reduce computing resource. During evolution search, this mechanism automatically adjusts the quantity of individuals and evolution frequency to adapt to the changing evolution state, which enormously reduces computing overhead.
\item We develop a constructive EA to ensure that DNN architectures are in near minimal scale for particular performance during evolution search. SANE construct DNN architectures by growing minimal structures gradually until they reach the required performance. It's beneficial to construct lightweight DNN architectures for devices with limited computing power. To protect evolution search from early convergence, a speciation scheme is set to maintain diversity in population by clustering population into different species.
\end{enumerate}

The rest of this paper is organized as follows. We review related works in Section \ref{sub:rw}. Next, we elaborate the theory and procedure of SANE in Section \ref{sub:sane}, and analyse its complexity in Section \ref{sub:ca}. Then, we conduct NAS experiments to evaluate SANE in Section \ref{sub:er}. Finally, we draw conclusions and suggest future work in Section \ref{sub:conclusion}.

\section{Related Work \label{sub:rw}}

In view of search strategy, NAS methods are mainly classified into reinforcement learning (RL)-based methods \cite{yesmina_rlr_2019} and EA-based ones \cite{stanley_designing_2019}. The latter is also referred to as neuroevolution \cite{floreano_neuroevolution_2008}. It has been reported that neuroevolution requires less computational time compared to RL-based NAS methods \cite{sun_completely_2020}. There are abundant mature neuroevolution theories lying idle subject to traditional simple DNN architecture composed by only full-connected nodes and shortage of computing resources in early age. Fortunately, neuroevolution studies boom again now, since new types of DNN architectures have emerged, such as Convolutional Neural Network (CNN) \cite{lecun_gradient-based_1998, khan_survey_2020}, Generative Adversarial Network (GAN) \cite{goodfellow_generative_2014, gonog_review_2019, Wang2022SEEM}, Long Short-Term Memory (LSTM) \cite{hochreiter_long_1997, van_houdt_review_2020}.

\begin{figure}[!htbp]
  \centering
  \subfigure[]{
    \label{fig:gss}
    \includegraphics[width=0.5\textwidth]{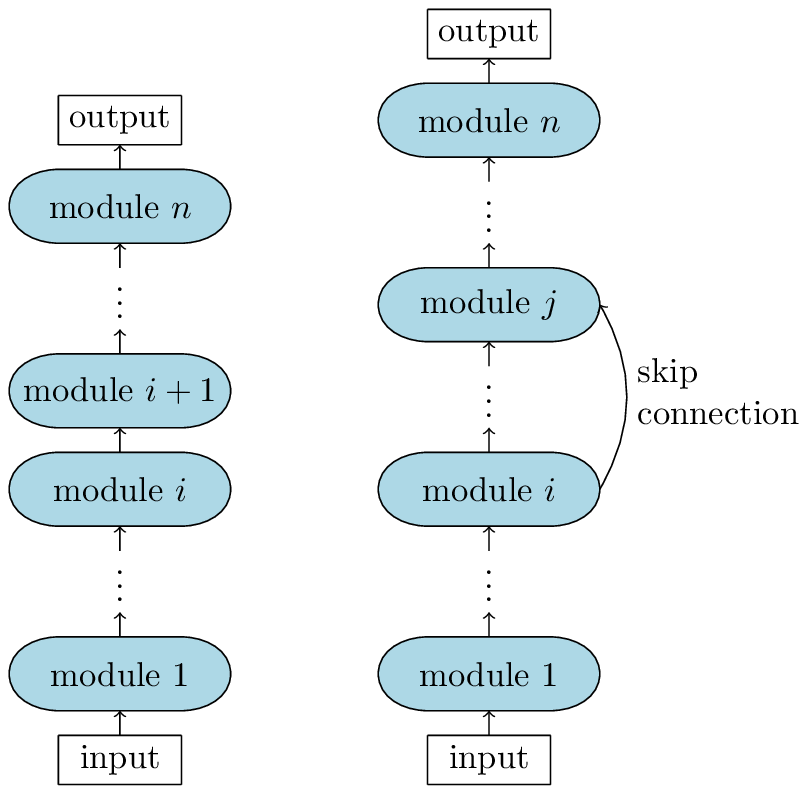}}
  \subfigure[]{
    \label{fig:css}
    \includegraphics[width=0.12\textwidth]{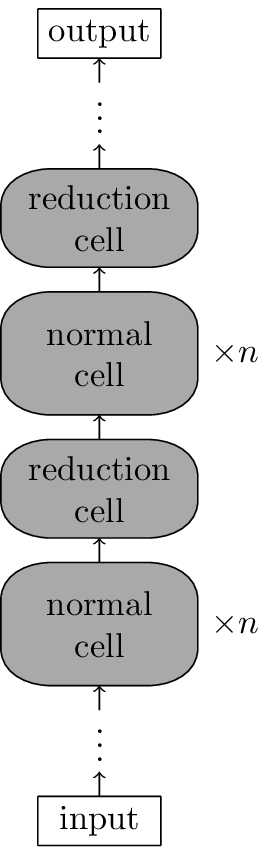}}
  \caption{Search Space: (a) chain-structured global search space without and with skip connection; (b) NASNet cell-based search space, $n$ normal cells followed by $1$ reduction cell.}
  \label{fig:SearchSpace}
\end{figure}

The two main study aspects in neuroevolution researches are search space and EA. Search space defines patterns of DNN architectures a neuroevolution method can theoretically discover. NAS-RL \cite{zoph_neural_2017} and MetaQNN \cite{baker_designing_2017} adopt global search space to generate chain-structured DNN architecture as shown in Figure~\ref{fig:gss}. Cell-based search space was firstly proposed in NASNet \cite{zoph_learning_2018}. Cell refers to fixed modules in many effective handcrafted DNN architectures. In cell-based search space, cells of the same type are integrated by several same modules in fixed topology as shown in Figure~\ref{fig:css}. Global and cell-based search space are widely used in NAS methods. In principle, global search space encompasses cell-based evolution space. It preserves relatively large search freedom degree while makes search require more computing resources. Cell-based evolution space provides more efficiency since it imposes more restriction on latent DNN architectures, while sacrifices search freedom degree in some extent.

EA, including Genetic Algorthims (GA) \cite{holland_adaptation_1992}, Genetic Programming (GP) \cite{koza_genetic_1992}, Evolution Strategy (ES) \cite{beyer_evolution_2002}, are almost identical in basic evolution steps. Neuroevolution researches mainly focus on implementation details of evolution procedure, especially the two most important issues: encoding scheme and evolution operations \cite{yao_evolving_1999}.

Since diversity is the key to make innovation possible \cite{gould_full_2011}, the evolution idea that randomly initialize, mutate and crossover individuals is widely adopted among neuroevolution methods \cite{liu_hierarchical_2018, real_regularized_2019}. This evolution idea leads to encoding schemes with genotype of fixed pattern naturally. For example, Genetic-CNN \cite{xie_genetic_2017} treats CNN architecture as a composition of three segments, the entire genotype is a fixed length binary string of the adjacency matrices for the three segments. Crossover swaps segments between selected pairs of individuals, mutation then randomly flips bits of the adjacency matrices. Obviously, this kind of encoding scheme and their associated evolution operations could not ensure obtained DNN architectures with relatively small scale. On the contrary, NeuroEvolution of Augmenting Topologies (NEAT) \cite{stanley_evolving_2002} adopts constructive evolution idea. It initializes phenotypes with minimal structure, then adds or grows genes to genotypes only when they could benefit the phenotypes during evolution, so that individuals are always in near-minimal scale for best fitness in the evolution generation. Without diversity in the initial population, evolution would be trapped into early convergence. NEAT solves this problem by designing a speciation scheme that allows individuals to compete primarily within their own species.

Selection as a relatively independent search operation pushes forward evolution process by retaining individuals with better fitness. Selection in Genetic-CNN \cite{xie_genetic_2017} adopts the idea of Russian roulette \cite{blickle_comparison_1996} that choose individuals with probabilities proportional to their fitness. Large-scale Evolution \cite{real_large-scale_2017} uses tournament selection \cite{goldberg_comparative_1991} that randomly chooses two individuals from population and compares their fitness, removes the worse one and selects the best one to duplicate as an offspring. These are two widely adopted selection operations, but not effective enough for fitness improvement of population.

\section{Self-Adaptive NeuroEvolution (SANE) Approach \label{sub:sane}}

We come up with a type-free search space to set evolution border for different types of DNN architectures by uniform definitions as detailed in Section \ref{sub:ss}. Based on type-free search space, we propose a constructive EA to evolve DNN architecture from minimal scale, in which a set of evolution operations are designed to add or modify genes in genotypes, as well as a speciation scheme is developed from NEAT to maintain diversity in population as described in Section \ref{sub:cces}. In addition, a evolution adjustment mechanism is brought up to adjust evolution exploration and exploitation setting according to the changing evolution search state as explained in Section \ref{sub:saam}. As for performance estimation, we import an early-stop mechanism to accelerate the estimation speed as exposited in Section \ref{sub:pe}.

\subsection{Type-free Search Space \label{sub:ss}}

The structure similarities among different DNN architectures reside in their composition, organization, and integration. Specifically, DNN architectures are composed by several particular kinds of modules, which are organized by a few regions with specific function. These modules and regions are integrated by connections in regular patterns. Based on the observation above, we define three concepts to formalize a type-free evolution space:

\begin{enumerate}[1)]
\item Cell $e$. We define cell as structure unit of DNN architectures. In SANE, one cell is composed by one core module with trainable parameters and some affiliated modules without trainable parameters. Core and affiliated modules are enabled to be added or modified, and latter can be removed. Cells of the same type are not constrained to keep identical to each other except the type of their core modules which determines their cell type.

\item Organ $o$. We treat regions of fixed functions in DNN architectures as organs composed by cells in specific types. For instance, a CNN architecture can be conceived to be integrated by one feature organ and one classifier organ.

\item Connection rule $r$. Connection rule assigns the quantity ceiling of in-cells and out-cells for each cell by connection degree $r_{d}$, and restricts the direction of data flow in connection by connection relations $r_{r}$.
\end{enumerate}

\begin{figure}[!htbp]
  \centering
  \includegraphics[width=0.6\textwidth]{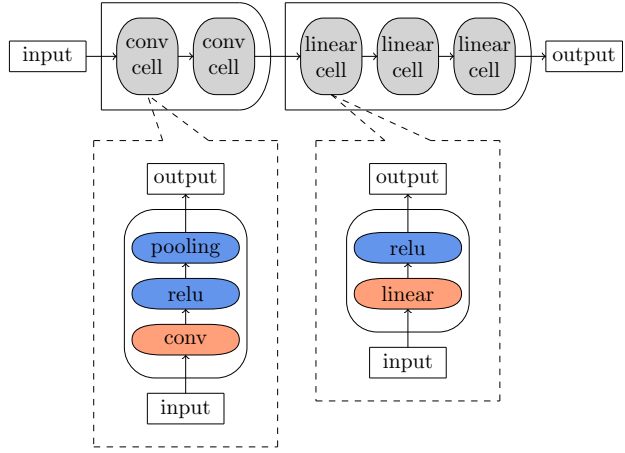}
  \caption{LeNet representation in type-free search space. LeNet architecture is composed by $2$ convolution modules, $3$ linear modules, as well as some activation modules and pooling modules. It is represented by $2$ convolution cells within feature organ and $3$ linear cells in classifier organ in type-free search space.}
  \label{fig:cellorgan}
\end{figure}

LeNet, for example, is represented by $5$ cells in $2$ organs in type-free search space as shown in Figure~\ref{fig:cellorgan}.

Specifically, a type-free evolution space is formulized as Eq.~\eqref{eq:searchspace}, where $\boldsymbol{E}$ is a cell set that declare valid cell types, $\mathbf{O}$ is an organ set preset with different cell subsets $\boldsymbol{E}^{(o)}\subseteq\boldsymbol{E}$ and $\boldsymbol{R}$ assigns connection degree and connection relations.

\begin{equation}
\mathbb{S}=
\begin{cases}
    \boldsymbol{E}=\{e_{i},e_{j},e_{k},\ldots\}, & \boldsymbol{E}^{(o)}\subseteq\boldsymbol{E};\\
    \boldsymbol{O}=\{o_{x},o_{y},\ldots\}, & \boldsymbol{E}^{(o_{x})}=\{e_{i},\ldots\},\boldsymbol{E}^{(o_{y})}=\{e_{j},\ldots\},\ldots;\\
    \boldsymbol{R}=\{r_{d},r_{r}\}, & r_{d}=n,r_{r}=\{(o_{x}, o_{y}),\ldots;(e_{i}, e_{j}),\ldots\}.
\end{cases}\label{eq:searchspace}
\end{equation}

\subsection{Cell-centered Constructive EA \label{sub:cces}}

In SANE, individuals are the same with a few necessary modules in initial evolution population. As evolution proceeds, individuals grow by evolution operations only when it improves their fitness.

\begin{figure}[!htbp]
  \centering
  \includegraphics[width=0.8\textwidth]{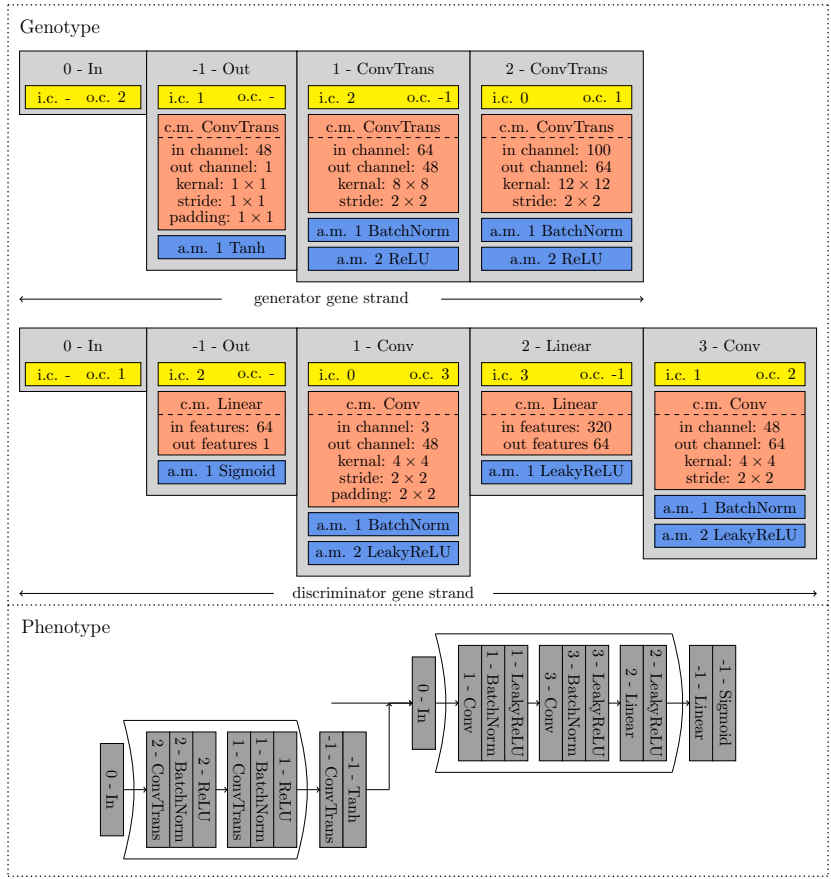}
  \caption{Genotype and phenotype of a GAN. The genotype contains two gene strands, one for generator organ, the other for discriminator organ. In cell genes, i.c. and o.c. indicate the cell key of in-cell and out-cell, c.m. and a.m. represent core module and affiliated module respectively. In phenotype, each module is attached with a cell key accordingly.}
  \label{fig:genopheno}
\end{figure}

We directly encode phenotypes into genotypes that contains several gene strands. Each cell gene strand records the attributions of cells for one organ, as well as the relation between cells. Figure~\ref{fig:genopheno} shows the genotype and phenotype of discriminator organ in a GAN which is composed by $2$ convolution cells and $1$ linear cell.

\begin{figure}[!htbp]
  \centering
  \subfigure[]{\includegraphics[width=0.8\textwidth]{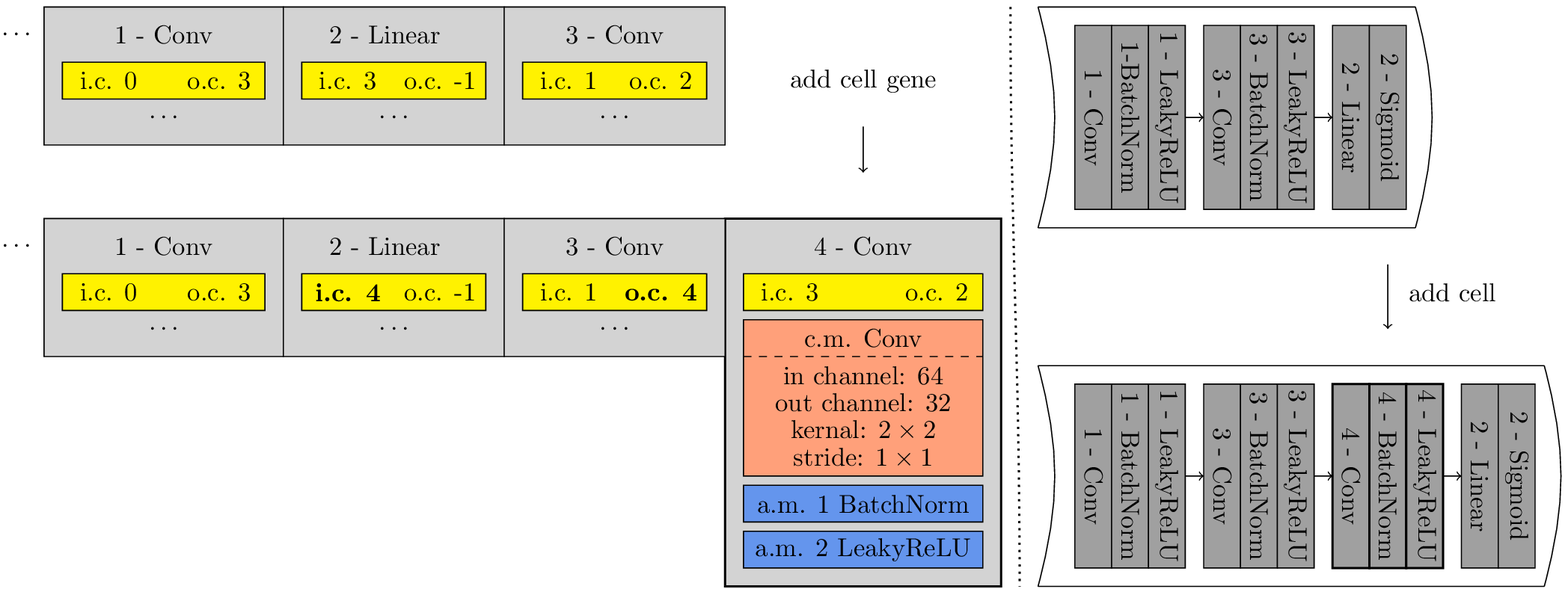}}
  \subfigure[]{\includegraphics[width=0.8\textwidth]{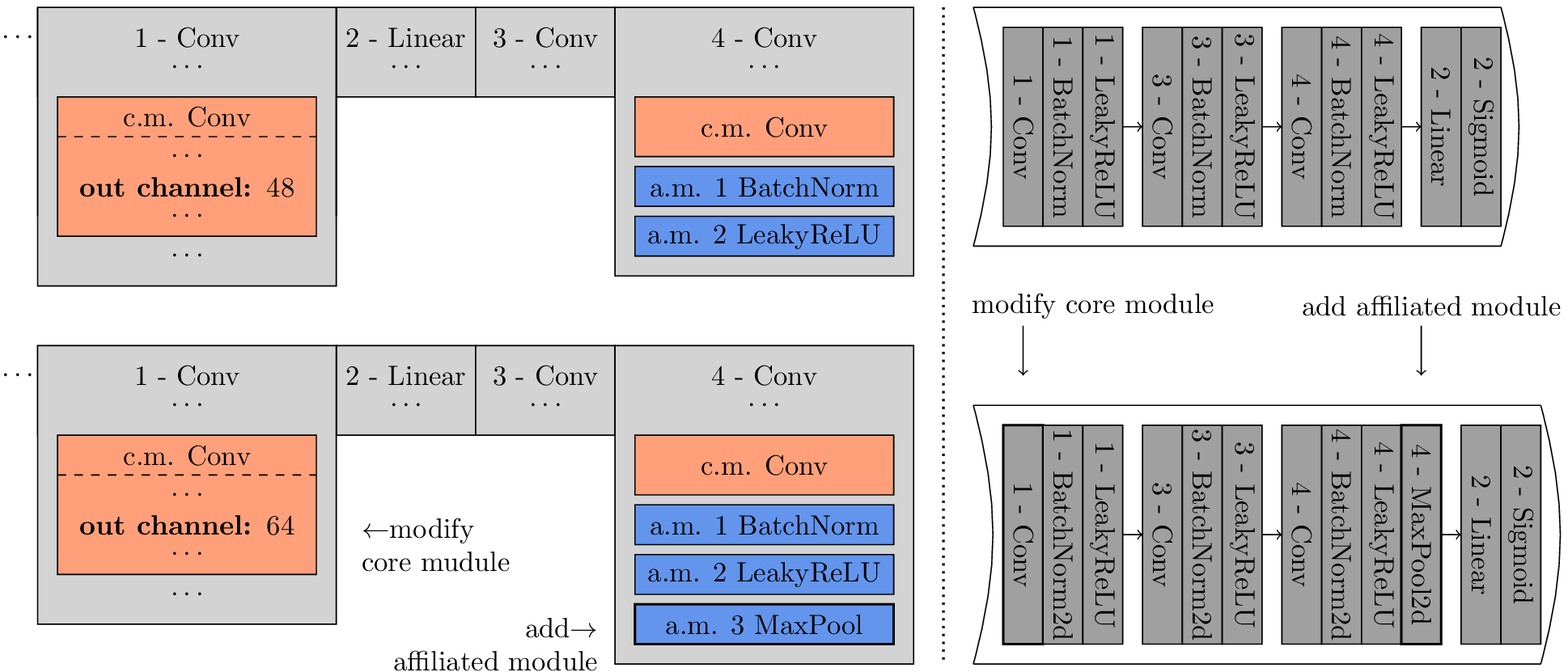}}
  \caption{Some mutation operations in SANE. (a) one case of adding cell, adding one cell gene to the tail of cell gene strand; (b) two cases of modifying cell, increasing kernel size of convolution module, and adding affiliated module to cell.}
  \label{fig:mutationoperations} 
\end{figure}

\newpage
Based on the cell-based encoding scheme, we design two kinds of cell-centered mutation operations $\mathbf{M}$ in different granularity: adding cell and modifying cell. The former adds one cell to an organ at random place within connection rule, while the later randomly chooses and modifies one attribution of a cell including the settings of core and affiliated modules. Figure~\ref{fig:mutationoperations} illustrates some specific mutation operations.

\begin{figure}[!htbp]
  \centering
  \includegraphics[width=0.8\textwidth]{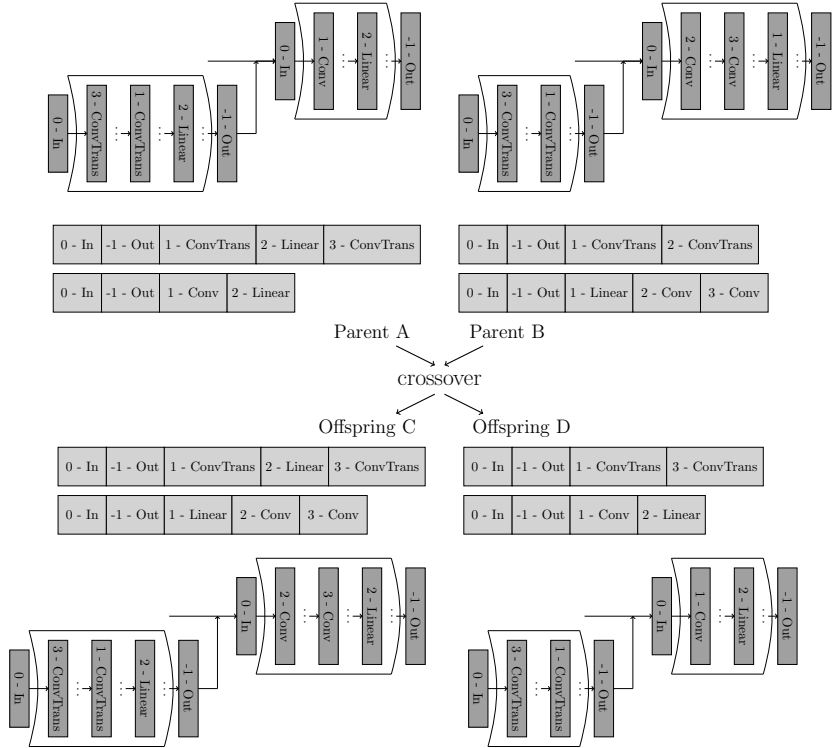}
  \caption{Crossover between two GAN genotypes. The crossover switch discriminator gene strands of parent A and B to produce two offsprings C and D. Their phenotypes are depicted accordingly.}
  \label{fig:crossover}
\end{figure}

\newpage
Besides, we design organ-centered crossover operation $\mathbf{C}$: switch two gene strands of the same organs from two parents. Figure~\ref{fig:crossover} illustrates crossover operation between two genotypes in GAN evolution, as well as their phenotypes.

\begin{algorithm}
\caption{Speciation\label{alg:Speciation}}

\SetKwInOut{Input}{input}
\SetKwInOut{Output}{output}
\SetKwData{None}{None}

\Input{genotypes $\mathnormal{G}$, existing species set $\mathnormal{S}$, compatibility distance ${\tau}_d$}
\Output{updated species set $\mathnormal{S}$}
	
\BlankLine

\For{species $s\in \mathnormal{S}$}{
	$d_t\leftarrow {\tau}_d$\;
	\For{genotype $g\in \mathnormal{G}$}{
		$d\leftarrow {\mathcal{D}}(g, g_s)$	\tcp*{$g_s$ is representative genotype of $s$}
		\If{$d<d_t$}{
			$d_t \leftarrow d$\;
			$g_t \leftarrow g$\;
    	}
	}
	\If{$g_t$}{
		$g_s\leftarrow g_t$\;
		$s\leftarrow \emptyset$\;
		$s\leftarrow s \cup \{g_s\}$\;
	}
}

\For{$g\in \mathnormal{G}$}{
	$s_t\leftarrow \emptyset$\;
	$d_t\leftarrow {\tau}_d$\;
	\For{$s\in \mathnormal{S}$}{
		$d\leftarrow \mathcal{D}(g, g_s)$\;
		\If{$d<d_t$}{
			$d_t \leftarrow d$\;
			$s_t = s$\;
    	}	
	}
	\If{$s_t \neq \emptyset$}{
		$s_t\leftarrow s_t \cup \{g\}$
	}
	\Else{
		$s_n\leftarrow \emptyset$\;
		$s_n\leftarrow s_n \cup \{g\}$\;
		$\mathnormal{S}\leftarrow \mathnormal{S} \cup \{s_n\}$\;
	}
}
\end{algorithm}

To protect diversity in population, we develop a speciation scheme that groups individuals into different species according to the compatibility distances between every two genotypes $g_{i}$ and $g_{j}$. The compatibility distance $\mathcal{D}(g_{i},g_{j})$ is measured by the summation of cell quantity gap as formulized in Eq.~\eqref{eq:distance}, where $n_{oe}$ represents the quantity gap of cell $e$ in organ $o$, $c_{oe}$ represents preset corresponding coefficient. Based on these compatibility distances and a preset compatibility distance threshold $\tau_{d}$, population is speciated by Algorithm~\ref{alg:Speciation}.

\begin{equation}
\mathcal{D}(g_{i},g_{j})=\sum_{o}\sum_{e}c_{oe}n_{oe}.\label{eq:distance}
\end{equation}

\subsection{Self-adaptive Search Adjustment Mechanism \label{sub:saam}}

Constructive search methods possess the following nature: at early search stage, small individuals search along relatively few directions and their fitness improves fast, and at late search stage, large individuals show a reverse trend. If constructive search methods could adjust search settings to adapt to the nature, it would save much computing resource and accelerate search speed. Based on this assumption, we design a self-adaptive search adjustment mechanism, which self-adaptively adjusts search settings in regard of exploring and exploiting type-free search space through two hyperparameters:

\begin{enumerate}[1)]
\item Mutation Time per Generation (abbreviated as TpG), $T$. In most neuroevolution methods, individuals are mutated once in each generation, while they are enabled to be mutated for several times indicated by TpG to exploit search space deeper in SANE. TpG is initialized by $\lambda_{T}$, and adjusted according to the change of best fitness $F$ in the population from last generation $k-1$ to current generation $k$ as formulated in Eq.~\eqref{eq:T}, where $\xi_{T}$ is the step value of TpG.

\begin{equation}
T^{(k+1)}=
\begin{cases}
    \lambda_{T}, & k=0~\text{or}~F^{(k)}>F^{(k-1)}~\text{or}~T^{(k)}>N^{(k)};\\
    T^{(k)}+\xi_{T}, & F^{(k)}\leqslant F^{(k-1)}~\text{and}~T^{(k)}\leqslant N^{(k)}.
\end{cases}\label{eq:T}
\end{equation}

\item Offspring Number per Individual (abbreviated as NpI), $N$. As search proceeds, more offsprings are in demand to cover increasing evolution directions emerging in population, so each individual is enabled to duplicate more offsprings indicated by NpI to explore search space wider when offsprings in current quantity don't improve the best fitness for several generations with different exploitation settings. NpI is initialized by $\lambda_{N}$, and adjusted depending on value of TpG as formulated in Eq.~\eqref{eq:N}, where $\xi_{N}$, $\tau_{N}$ is step value and threshold of NpI.

\begin{equation}
N^{(k+1)}=
\begin{cases}
    \lambda_{N}, & k=0;\\
    N^{(k)}, & T^{(k)}\leqslant N^{(k)};\\
    N^{(k)}+\xi_{N}, & T^{(k)}>N^{(k)}~\text{and}~N^{(k)}\leqslant\tau_{N}.
\end{cases}\label{eq:N}
\end{equation}

\end{enumerate}

Self-adaptive adjustment mechanism adjusts exploitation and exploration setting for next generation at the end of current evolution generation. It firstly increases exploitation depth by augment TpG if the best fitness doesn't improve. When TpG becomes bigger than NpI in current generation, it then increases exploration width by augment NpI and reset exploitation depth with initial TpG.

\subsection{Performance Estimation and Selection \label{sub:pe}}

\begin{figure}[!htbp]
  \centering
  \subfigure[]{\includegraphics[width=0.45\textwidth]{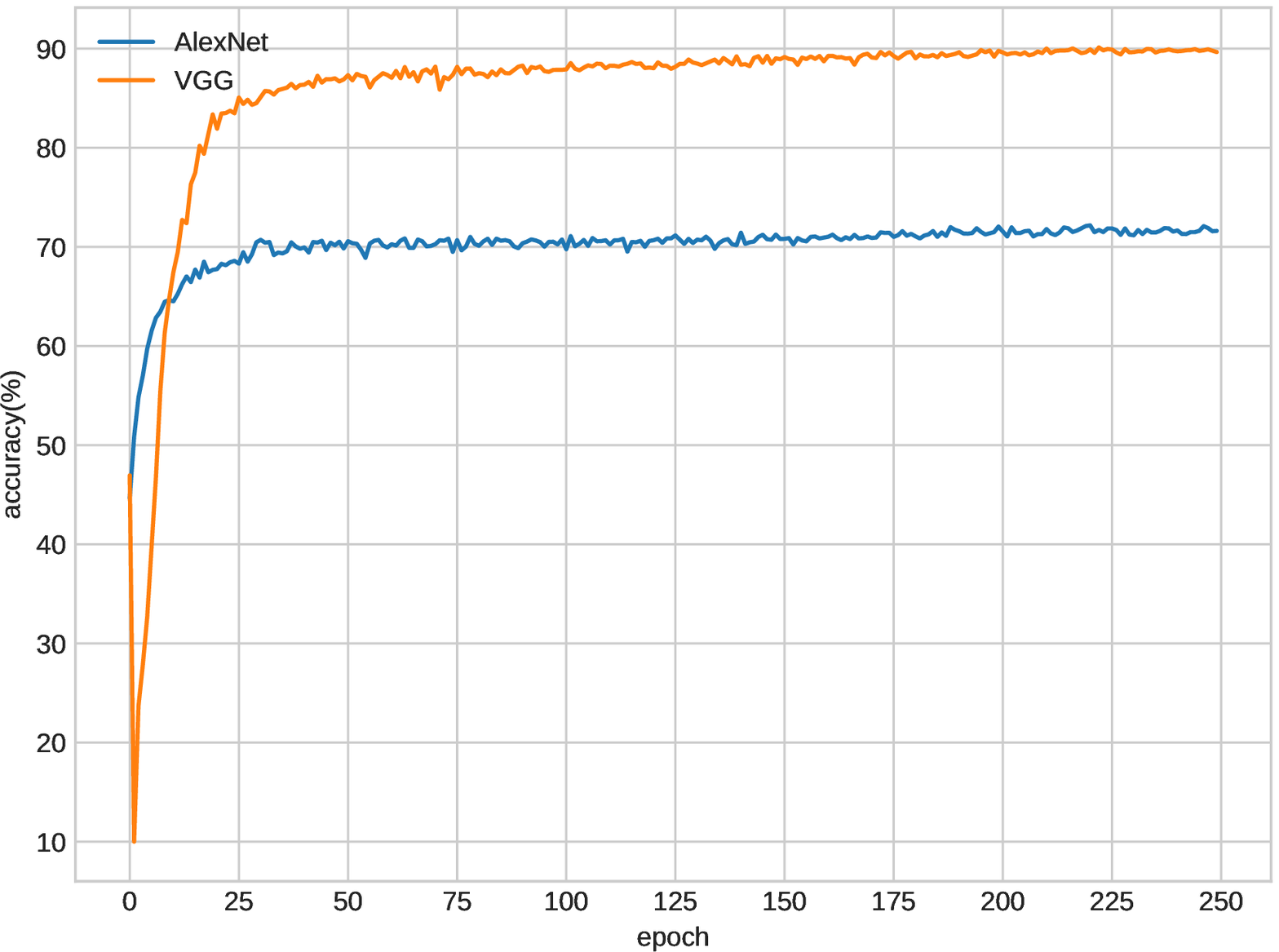}}
  \subfigure[]{\includegraphics[width=0.45\textwidth]{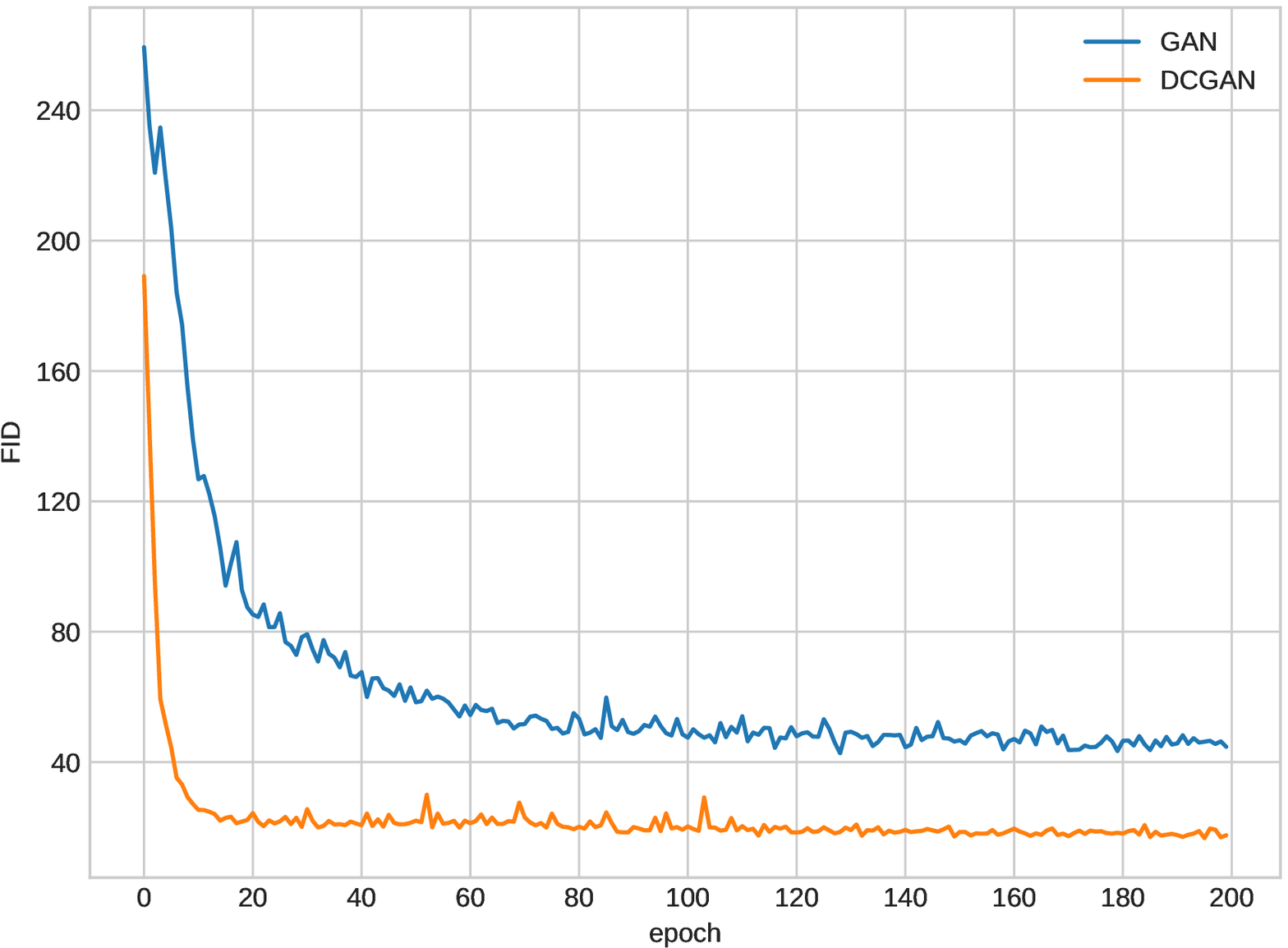}}
  \caption{Different DNN training process. (a) The accuracy change of AlexNet and VGG during training process for CIFAR10. (b) The FID change of GAN and DCGAN during training porcess for MNIST.}
  \label{fig:est}
\end{figure}

In SANE, performance estimation occupies most time and computation resource during evolution, since we estimate individuals' fitness by training and validating their phenotypes in a given task. To relieve the burden, we design an early-stop training mechanism based on an observation that DNNs would soon tend to convergence after being trained for a few epochs as shown in Figure~\ref{fig:est}.

\begin{algorithm}
\caption{SANE\label{alg:SANE}}

\SetKwInOut{Input}{input}
\SetKwInOut{Output}{output}

\SetKwFunction{Speciation}{Speciation}
\SetKwFunction{Range}{range}

\Input{individuals $\mathnormal{I}$ in initial population, evolution settings}
\Output{individual $o_b$ with best fitness}
	
\For{$k \leftarrow \{1, 2, \cdots, {\tau}_k\}$}{
	offsprings $\mathnormal{O}\leftarrow \emptyset$\;
	\For{$n$ in $N$}{
		$\mathnormal{O} \leftarrow \mathnormal{O} \cup \mathnormal{I}$\;
	}
	\For{$t$ in $T$}{	
		$\mathbf{M} (\mathnormal{G}_o, p_m)$\;
		$\mathbf{C} (\mathnormal{G}_m, p_c)$
		\tcp*{$p_m$, $p_c$ is preset mutation and crossover probability}
	}
	$\mathnormal{I} \leftarrow \mathnormal{I} \cup \mathnormal{O}$\;
	$\mathnormal{S}_n \leftarrow$ \Speciation{$\mathnormal{G}_I, \mathnormal{S}, {\tau}_d$}\;
	$F^{(k)} \leftarrow \mathbf{T}_i (\mathnormal{P}_I, t_i)$	
	\tcp*{$\mathbf{T}_i$ is incomplete training}
	\If{$F^{(k)} > F_b$}{	
		$F_b \leftarrow F^{(k)}$
		\tcp*{$F_b$ is best fitness in current generation}
		\If{$F_b > {\tau}_F$}{
			$F_c \leftarrow \mathbf{T}_c (p_b, t_c)$
			\tcp*{$\mathbf{T}_c$ is complete training}
			\If{$F_c > {\tau}_{F_c}$}{
				\Return{$o_b$}\;
			}
		}
		$T \leftarrow {\lambda}_{T}$\;
	}
	\Else{
		\If{$T > N$}{
			$N \leftarrow N+{\xi}_N$\;
			$T \leftarrow {\lambda}_{T}$\;		
		}
		\Else{
			$T \leftarrow T+{\xi}_T$\;
		}
	}
	$\mathnormal{I} \leftarrow \emptyset$\;
	\For{$s \in \mathnormal{S}_n$}{
		$q \leftarrow \mathcal{Q} (s)$\;
		$I_s \leftarrow \mathbf{S} (s, q)$\;
		$\mathnormal{I} \leftarrow \mathnormal{I} \cup I_s$\;
	}
}
\end{algorithm}

To be specific, the idea is as follows: incompletely train all individuals within a few epochs $t_{i}$ which indicate the `early-stop' to get their incomplete fitness $F$ and verify whether their incomplete fitness $F$ reach the incomplete fitness threshold $\tau_{F}$. For any individual, if $F>\tau_{F}$, it will be completely trained within many more epochs $t_{c}$ to obtain its complete fitness $F_{c}$, if the complete fitness reaches the required fitness threshold $F_{c}>\tau_{F_{c}}$ for the particular task, SANE will record it as satisfied individual and output its phenotype. The incomplete fitness threshold $\tau_{F}$, complete fitness threshold $F_{c}>\tau_{F_{c}}$, incomplete training epoch $t_{i}$, complete training epoch $t_{c}$ are preset parameters.

After estimation, selection operation $\mathbf{S}$ chooses some members with higher fitness from each species to form new population for next generation. The quantity of selected members in species $s_{i}$ is calculated by Eq.~\eqref{eq:Q}, where $q_{i}$ is the quantity of members in species $s_{i}$, and $\tau_{q}$ is the preset quantity ceiling of whole population.

\begin{equation}
\mathcal{Q}(s_{i})=
\begin{cases}
    q_{i}, & \sum_{i}q_{i}\leqslant\tau_{q};\\
    \frac{q_{i}}{\sum_{i}q_{i}}\tau_{q}, & \sum_{i}q_{i}>\tau_{q}.
\end{cases}\label{eq:Q}
\end{equation}

The whole process of SANE is described in Algorithm~\ref{alg:SANE}. In each generation, all individuals $\mathnormal{I}$ are firstly duplicated as offsprings $\mathnormal{O}$, and all offspring's genotypes $\mathnormal{G}_{o}$ are mutated and crossover with preset probabilities. The operated genotypes and original individuals $\mathnormal{I}$ are then speciated into different species. Next, some individuals are sampled from each species to decoded into phenotypes $\mathnormal{P}_I$ to estimate incomplete fitness. Subsequently, self-adaptive adjustment mechanism adjusts the exploitation and exploration setting according to the best fitness change. At last, some members in each species are selected to form new population for next generation.

\section{Complexity Analysis \label{sub:ca}}

Before evolution search with SANE, we set quantity ceiling $c_{e_{i}}$ for each cell type $e_{i}$ in DNN architectures. Suppose that there are $2^{l_{e_{i}}}$ states for each cell type, different states mean that cells are in the same type but with different attributions. For arbitrary cell $e_{i}$, it can be represented by a state string $x_{e}=(b_{1}\cdots b_{i}\cdots b_{l_{e_{i}}})$ where $b_{i}\in\{0,1\}$. Therefore, each DNN architecture in type-free evolution space can be represented by a binary string $x=(b_{1}\cdots b_{Z})$ where $Z=\sum_{i=1}c_{e_{i}}l_{e_{i}}$.

Suppose there is $1$ unique optimal DNN architecture in type-free search space whose binary string representation is $x^{*}=(1\cdots1)$, and the number of $1$ bits in other DNN architectures' binary strings are positively associated with their fitness, then the neuroevolution problem in this study can be turned into a subset problem solved by EA like that in \cite{sami_optim_1994}.

Let $d(x)$ be the distance between $x$ and $x^{*}$, and given a population $X=\{x_{1},\cdots,x_{n}\}$, then
\begin{equation}
d(X)=\min\{d(x):x\in X\},
\label{eq:dp}
\end{equation}
which is used to measure the distance of a population to the optimal DNN architecture.

As evolution search goes on, $x$ will drift towards $x^{*}$ and hit it eventually. Define the stopping time of an EA as $\tau=\min\{k:d(X_{k})=0\}$, which is the first hitting time on $x^{*}$, then we can analyse the average time complexity of EA through estimating the expected first hitting time $H[\tau]$.

According to the Theorem 5 in \cite{he_drift_2001}: given the family of subset sum problems and the EA to solve them, for any initial population $X$ with $d(X)>0$, we have
\begin{equation}
H[\tau]\leqslant f(Z),
\label{eq:E_tau}
\end{equation}
where $f(Z)$ is a polynomial of $n$, $f(Z)=O(Z^{2})$.

\section{Experiments and Results \label{sub:er}}

To investigate SANE, we conducted CNN, GAN and LSTM NAS experiments on some CV tasks to valid its self-adaption in regard to DNN type and evolution state, as well as verify that it is able to construct DNN architectures with near-minimal scale for particular performance. In order to reduce computing consumption as much as possible and make experiments accessible, we chose classical DNN architectures as competitors and evolved new DNN architectures with the same components for benchmark tasks with standard datasets.

SANE is implemented by widely used deep learning development tool PyTorch \cite{paszke_pytorch_2019} in a workstation equipped with Ubuntu 20.04 operation system and Nvidia GeForce RTX 3090 graphics card ($24$ GB), and no evolution experiment spent more than $2.5$day/GPU, i.e., $60$h/GPU. 

\subsection{CNN Evolution}

CIFAR10 and CIFAR100 are two widely used benchmark datasets in CV tasks, each of which has $50,000$ training images and $10,000$ test images. The object in each image has different resolution, mixes with background, and occupies different position, which generally increases the difficulty of CV tasks. The difference between them is that CIFAR10 is a $10$-class classification, while CIFAR100 is a $100$-class classification. CNNs are typically estimated by performing image classification task in CIFAR10 and CIFAR100 datasets, so we adopt the classification accuracy (\%) as fitness of DNN architecture during evolution search.

\begin{equation}
\mathbb{S}_{CNN}=
\begin{cases}
    \boldsymbol{E}=\{e_{c},e_{l}\}, & \boldsymbol{E}^{(o)}\subseteq\boldsymbol{E};\\
    \boldsymbol{O}=\{o_{f},o_{c}\}, & \boldsymbol{E}^{(o_{f})}=\{e_{c}\},\boldsymbol{E}^{(o_{c})}=\{e_{l}\};\\
    \boldsymbol{R}=\{r_{d},r_{r}\}, & r_{d}=1,r_{r}=\{(o_{f}, o_{c});(e_{c}, e_{c}),(e_{l}, e_{l})\}.
\end{cases}\label{eq:scnn}
\end{equation}

We chose two representative CNN architectures, i.e. AlexNet \cite{krizhevsky_imagenet_2017} and VGG \cite{simonyan_very_2015}, as CNN competitors. These two CNNs contain two organs, i.e., feature and classifier, and are both composed by convolution and linear cell, so we set one uniform type-free search space for them as Eq.~\eqref{eq:scnn}, where $e_{c}$ and $e_{l}$ represent convolution and linear cell, $o_{f}$ and $o_{c}$ represent feature and classifier organ, respectively. Table~\ref{tab:cnnevo} generalizes the key evolution configuration in regard to CNN attributions, training, and evolution. The DNN config part specifies cell types in organs, for example, \texttt{\{conv:[[16/32,3,1,0], [batchnorm, relu, maxpool]]\}} indicates that a initial convolution cell contains a core module with $16$ (for AlexNet) or $32$ (for VGG) out channels, $3\times3$ kernel size, $1$ stride and $0$ padding, as well as batch normalization, rectified linear unit (relu)  \cite{nair_rectified_2010} and max pooling affiliated modules. The evolution config part sets the initial and ceiling individual quantity in population, the initial and step size of NpI and TpG, species limit, and distance threshold. Besides, some evolution probability hyper parameters are also given here, for instance, \texttt{conv attr prob:[40\%,15\%,15\%,15\%,15\%]} controls mutation probability of convolution module attributions. The training config determines the training ratio in each species, training loss function Cross Entropy \cite{murphy_machine_2012} and optimizer Adam \cite{kingma_adam_2015}.

\begin{table}[!ht]
\caption{CNN Evolution Configurations\label{tab:cnnevo}}
\centering
\begin{tabular}{ |ll| } 
\hline
\multicolumn{2}{|c|}{DNN config} \\ 
\hline
\texttt{DNN type:CNN} & \texttt{organ types:[feature,classifier]} \\ 
\texttt{feature cell types:} &  \\
\multicolumn{2}{|l|}{\texttt{\{conv:[[16/32,3,1,0],[batchnorm,relu,maxpool]]\}}} \\
\texttt{classifier cell types:} & \\
\multicolumn{2}{|l|}{\texttt{\{linear:[32/64,[relu]]\}}} \\
\hline
\hline
\multicolumn{2}{|c|}{evolution config} \\
\hline
\texttt{individual init:20} & \texttt{individual limit:50} \\
\texttt{NpI init:1} & \texttt{NpI step:1} \\
\texttt{NpI limit:10} & \\
\texttt{TpG init:1} & \texttt{TpG step:1} \\
\texttt{organ prob:[60\%,40\%]} & \texttt{add cell prob: 25\%} \\
\texttt{modify cell prob:50\%} & \texttt{crossover prob:25\%} \\
\multicolumn{2}{|l|}{\texttt{conv attr prob:[40\%,15\%,15\%,15\%,15\%]}} \\
\multicolumn{2}{|l|}{\texttt{conv attr growth factor:[8/32,2,2,2]}} \\
\multicolumn{2}{|l|}{\texttt{linear attr growth factor:16/64}} \\
\multicolumn{2}{|l|}{\texttt{species num limit:10/15}} \\
\multicolumn{2}{|l|}{\texttt{species distance threshold:1.0}} \\
\hline
\hline
\multicolumn{2}{|c|}{training config} \\
\hline
\texttt{train rate:50\%} &  \\
\texttt{incomplete train epochs:10} & \texttt{complete train epochs:250} \\
\texttt{train batches:128} & \texttt{learning rate:0.001} \\
\texttt{loss function:Cross Entropy} & \texttt{optimizer:Adam} \\
\hline
\end{tabular}
\end{table}

\begin{figure}
\centering
\includegraphics[width=1.0\linewidth]{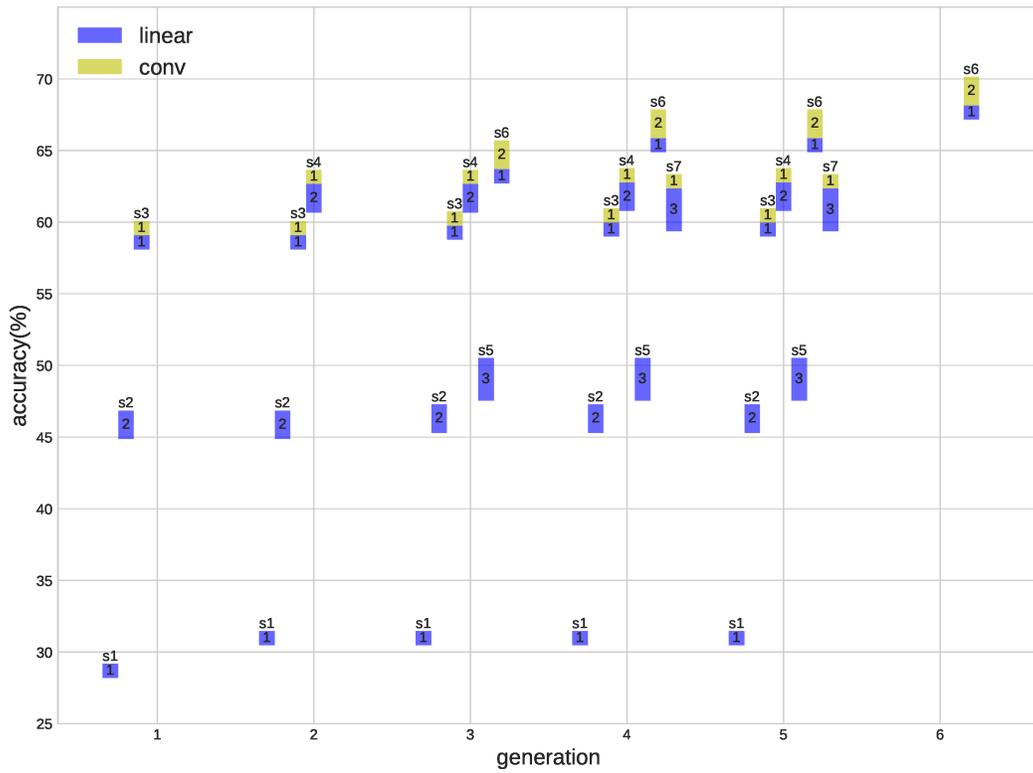}
\caption{The distribution of species incomplete fitness during SANE-Alex-e1 evolution process. The bottom line of each bar points the incomplete fitness of the species.}
\label{fig:alexevo}
\end{figure}

We firstly conducted AlexNet-type CNN architecture evolution experiment with CIFAR10 for $3$ times, and generated $3$ edition architecture, i.e., SANE-Alex-e1, SANE-Alex-e2 and SANE-Alex-e3 as detailed in Table~\ref{tab:cnnc10}. Taking SANE-Alex-e1 as example, the evolution process went through $6$ generations, during which $7$ species emerged. The species incomplete fitness is shown in Figure~\ref{fig:alexevo}. The best individual firstly emerges in generation $3$, species $6$. Its incomplete fitness is $62.70$\% at the beginning and reaches incomplete fitness threshold $65.37$\% after $3$ evolution generations.

We then conducted VGG-type CNN architecture evolution experiment with CIFAR10 for $3$ times, and generated $3$ edition architecture, i.e., SANE-VGG-C10-e1, SANE-VGG-C10-e2 and SANE-VGG-C10-e3 as detailed in Table~\ref{tab:cnnc10}. In VGG, the convolution cell contains an extra module, i.e. batch normalization, compared with that in AlexNet, while the linear cell is the same with that in AlexNet. The key search settings are basically the same with that in AlexNet-type CNN architecture evolution. 

\begin{figure}
\centering
\includegraphics[width=1.0\textwidth]{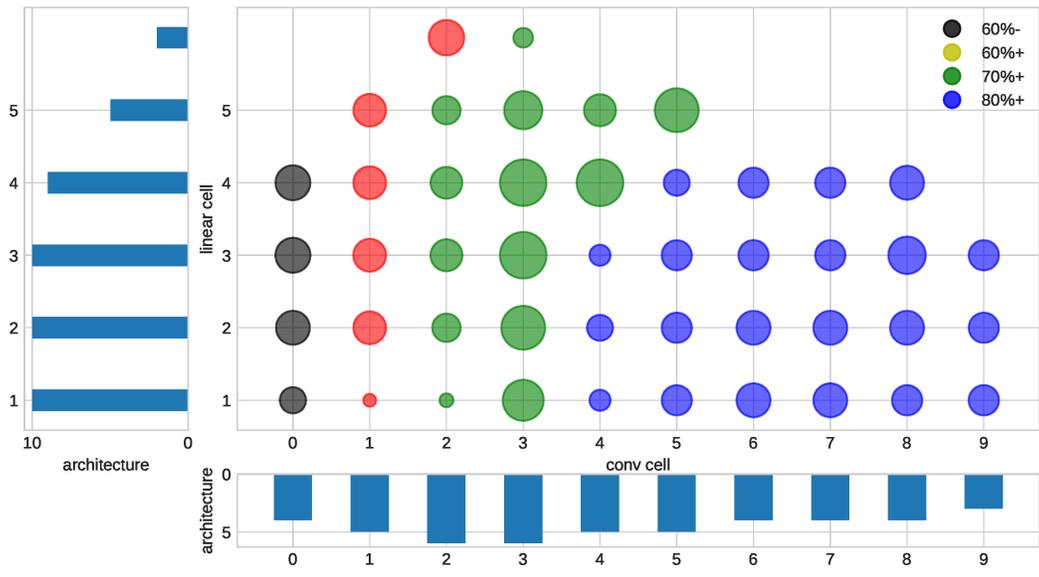}
\caption{The best architectures and their incomplete fitness in each species during SANE-VGG-C10-e1 evolution. In the main subplot (upper-right), the x-axis and y-axis show the number of convolution cells and linear cells in CNN architectures, the dots with 4 different colors represent different CNN architectures with different fitness stage range from $60$\%- (less than $60$\%) to $80$\%+ (more than $80$\%), while the bigger dots represent fitter CNN architectures among dots with the same color. The affiliated subplots (upper-left and lower-right) show the distribution of linear cell and convolution cell number in CNN architectures depicted in the main subplot. }
\label{fig:vggevo}
\end{figure}

Taking SANE-VGG-C10-e1 as example, the evolution went through $34$ generations, more than $46$ species survived during the evolution process. The distribution of the best CNN architectures and their incomplete fitness in all species emerged during evolution are illustrated in Figure~\ref{fig:vggevo}. It shows that as linear cell increasing (from bottom to top), there are CNN architectures in $4$ incomplete fitness stages. And as convolution cell increasing (from left to right), the fitness of CNN architectures increases gradually from $60$\%- to $80$\%+. CNN architectures with incomplete fitness above $80$\% are clustered together in right area of the main subplot, and no CNN architecture with more than $6$ linear cells survived during evolution. It's known that convolution has greater impact on the performance of CNN architectures, and the evolution process shows that SANE is capable to self-adaptively evolve DNN architectures in fitter directions. Moreover, the best CNN architecture is composed by $8$ convolution cells and $3$ linear cells, which indicates simply adding modules doesn't always lead to fitter CNN architectures. It verifies SANE is useful to find out the near-minimal CNN architectures for a given performance standard. Besides, the even distribution of linear cell and convolution cell quantities implies the diversity in population during evolution search.

\begin{table}[!ht]
\caption{Scale and performance of CNN architectures for CIFAR10 \label{tab:cnnc10}}
\centering{}
\begin{tabular*}{1\textwidth}{@{\extracolsep{\fill}}ccccccc}
\toprule
{\scriptsize{}method} & {\scriptsize{}cell} & {\scriptsize{}layer} & {\scriptsize{}parameter} & {\scriptsize{}acc(\%)} & {\scriptsize{}gen} & {\scriptsize{}time}  \tabularnewline  
\midrule
{\scriptsize{}AlexNet \cite{krizhevsky_imagenet_2017}} & {\scriptsize{}5 conv + 1 linear} & {\scriptsize{}15} & {\scriptsize{}2.47M} & {\scriptsize{}72.51} & - & - \tabularnewline
{\scriptsize{}SANE-Alex-e1 (ours)} & {\scriptsize{}2 conv + 1 linear} & {\scriptsize{}8} & {\scriptsize{}\textbf{41.17K}} & {\scriptsize{}\textbf{78.58}} & {\scriptsize{}6} & {\scriptsize{}37.68min/GPU} \tabularnewline
{\scriptsize{}SANE-Alex-e2 (ours)} & {\scriptsize{}2 conv + 1 linear} & {\scriptsize{}8} & {\scriptsize{}\textbf{26.87K}} & {\scriptsize{}\textbf{75.82}} & {\scriptsize{}6} & {\scriptsize{}36.88min/GPU} \tabularnewline
{\scriptsize{}SANE-Alex-e3 (ours)} & {\scriptsize{}2 conv + 1 linear} & {\scriptsize{}8} & {\scriptsize{}\textbf{33.55K}} & {\scriptsize{}\textbf{76.07}} & {\scriptsize{}5} & {\scriptsize{}30.03min/GPU} \tabularnewline
\midrule
{\scriptsize{}VGG17 \cite{simonyan_very_2015}} & {\scriptsize{}16 conv + 1 linear} & {\scriptsize{}55} & {\scriptsize{}20.04M} & {\scriptsize{}91.51} & - & - \tabularnewline
{\scriptsize{}SANE-VGG-C10-e1 (ours)} & {\scriptsize{}8 conv + 3 linear} & {\scriptsize{}32} & {\scriptsize{}\textbf{1.35M}} & {\scriptsize{}\textbf{91.80}} & {\scriptsize{}34} & {\scriptsize{}19.44h/GPU} \tabularnewline
{\scriptsize{}SANE-VGG-C10-e2 (ours)} & {\scriptsize{}7 conv + 1 linear} & {\scriptsize{}27} & {\scriptsize{}\textbf{2.55M}} & {\scriptsize{}\textbf{91.69}} & {\scriptsize{}28} & {\scriptsize{}23.77h/GPU} \tabularnewline
{\scriptsize{}SANE-VGG-C10-e3 (ours)} & {\scriptsize{}6 conv + 2 linear} & {\scriptsize{}25} & {\scriptsize{}\textbf{2.79M}} & {\scriptsize{}\textbf{91.73}} & {\scriptsize{}29} & {\scriptsize{}26.46h/GPU} \tabularnewline
\bottomrule
\end{tabular*}
\end{table}

\begin{table}[!ht]
\caption{Scale and performance of CNN Architectures for CIFAR100\label{tab:cnnc100}}
\centering
\begin{tabular*}{1\textwidth}{@{\extracolsep{\fill}}ccccccc}
\toprule
{\scriptsize{}method} & {\scriptsize{}cell} & {\scriptsize{}layer} & {\scriptsize{}parameter} & {\scriptsize{}acc(\%)} & {\scriptsize{}gen} & {\scriptsize{}time}  \tabularnewline 
\midrule
{\scriptsize{}VGG17 \cite{simonyan_very_2015}} & {\scriptsize{}16 conv + 1 linear} & {\scriptsize{}55} & {\scriptsize{}20.09M} & {\scriptsize{}63.86} & - & - \tabularnewline
{\scriptsize{}SANE-VGG-C100-e1 (ours)} & {\scriptsize{}7 conv + 1 linear} & {\scriptsize{}26} & {\scriptsize{}\textbf{1.29M}} & {\scriptsize{}\textbf{64.01}} & {\scriptsize{}19} & {\scriptsize{}38.81h/GPU} \tabularnewline
{\scriptsize{}SANE-VGG-C100-e2 (ours)} & {\scriptsize{}6 conv + 1 linear} & {\scriptsize{}24} & {\scriptsize{}\textbf{0.80M}} & {\scriptsize{}\textbf{65.98}} & {\scriptsize{}17} & {\scriptsize{}35.65h/GPU} \tabularnewline
{\scriptsize{}SANE-VGG-C100-e3 (ours)} & {\scriptsize{}7 conv + 1 linear} & {\scriptsize{}24} & {\scriptsize{}\textbf{2.19M}} & {\scriptsize{}\textbf{64.08}} & {\scriptsize{}23} & {\scriptsize{}49.08h/GPU} \tabularnewline
\bottomrule
\end{tabular*}
\end{table}

With the same evolution configuration, we conducted VGG-type CNN architecture experiment for CIFAR100. Both the scale and performance of CNN by SANE and their competitors for CIFAR10 and CIFAR100 are exhibited in Tables~\ref{tab:cnnc10} and \ref{tab:cnnc100}, respectively. We recorded the total cell number (cell), total trainable parameter number (parameter), and the image classification accuracy (acc) of CNN architectures. Besides, we recorded the evolution generation (gen) and time of each experiment. The results show that with identical CNN components, CNNs by SANE could reach similar accuracy compared to its competitor, while contain fewer cells and trainable parameters.

\subsection{GAN Evolution}

In SANE, we adopted Frechet inception distance (FID) \cite{heusel_gans_2017} to estimate the performance of GAN architecture. FID measures the distance between feature vectors of real and generated images, which indicates the similarity in terms of statistics on features of raw and generated images extracted by Inception-v3 model \cite{szegedy_rethinking_2016}. The lower the score is, the more similar the real and generated images are. To estimate GAN architectures emerged during evolution, we chose $1000$ raw images from specific dataset, and calculated the FID between these raw images and $1000$ images generated by each GAN architecture. For each GAN by SANE, its fitness is inversely proportional to its FID.

We chose MNIST and FashionMNIST as training dataset and raw images. MINST is a dataset of handwritten digit images, while FashionMNIST is a dataset of clothes images. Both of them contain $60,000$ training and $10,000$ test gray-scale images. Furthermore, the digits and clothes in the image have been size-normalized and centered in a fixed-size which is beneficial for generative model to generalize their features.

We chose the widely used DCGAN \cite{radforc_unsupervised_2016} as a competitor. Most GANs contain two organs, i.e. generator organ and discriminator organ. DCGAN is composed by convolution cell and convolution transpose cell, so we set the type-free search space as Eq.~\eqref{eq:sgnn}. Cell $e_{t}$ represent convolution transpose cell, organ $o_{g}$ and $o_{d}$ represent generator and discriminator organ respectively, connection degree $r_{d}$ indicates that each cell connects $1$ input cell and $1$ output cell at most, connection relation $r_{r}$ constrains data flow from generator organ to discriminator organ, while each of them is enable to receive and output data independently.

\begin{equation}
\mathbb{S}_{GAN}=
\begin{cases}
    \boldsymbol{E}=\{e_{t},e_{c}\}, & \boldsymbol{E}^{(o)}\subseteq\boldsymbol{E};\\
    \boldsymbol{O}=\{o_{g},o_{d}\}, & \boldsymbol{E}^{(o_{g})}=\{e_{t}\},\boldsymbol{E}^{(o_{d})}=\{e_{c}\};\\
    \boldsymbol{R}=\{r_{d},r_{r}\}, & r_{d}=1,r_{r}=\{(o_{g}, o_{d}),o_{g},o_{d};(e_{t}, e_{t}),(e_{c}, e_{c})\}.
\end{cases}\label{eq:sgnn}
\end{equation}

Table~\ref{tab:ganevo} generalizes key evolution configuration in regard to CNN attributions, training and evolution. Most configurations are the same with those in CNN evolution. In generator organ, convolution transpose cell (convtrans) contains a convolution transpose module \cite{dumoulin_trans_2016} which is able to increase the spatial dimensions of intermediate feature maps. In discriminator organ, convolution cell contains a leaky rectified linear unit (leakyrelu) \cite{xu_leakyrelu_2015} activation module. The loss function is Binary Cross Entropy (BCE) \cite{murphy_machine_2012}.

\begin{table}[!ht]
\caption{GAN Evolution Configurations\label{tab:ganevo}}
\centering
\begin{tabular}{ |ll| } 
\hline
\multicolumn{2}{|c|}{DNN config} \\ 
\hline
\texttt{DNN type:GAN} & \texttt{organ types:[generator,discriminator]} \\ 
\texttt{generator cell types:} &  \\
\multicolumn{2}{|l|}{\texttt{\{convtans:[[32,2,1,0],[batchnorm,relu]]\}}} \\
\texttt{discriminator cell types:} & \\
\multicolumn{2}{|l|}{\texttt{\{conv:[32,[[32,2,1,0],[batchnorm,leakyrelu]]\}}} \\
\hline
\hline
\multicolumn{2}{|c|}{evolution config} \\
\hline
\texttt{organ prob:[50\%,50\%]} & \\
\multicolumn{2}{|l|}{\texttt{conv attr prob:[40\%,20\%,20\%,20\%]}} \\
\multicolumn{2}{|l|}{\texttt{conv attr growth factor:[8,2,1,1]}} \\
\multicolumn{2}{|l|}{\texttt{convtrans attr prob:[40\%,20\%,20\%,20\%]}} \\
\multicolumn{2}{|l|}{\texttt{convtrans attr growth factor:[8,2,1,1]}} \\
\multicolumn{2}{|l|}{\texttt{species num limit:10}} \\
\multicolumn{2}{|l|}{\texttt{species distance threshold:1.0}} \\
\hline
\hline
\multicolumn{2}{|c|}{training config} \\
\hline
\texttt{incomplete train epochs:10} & \texttt{complete train epochs:200} \\
\texttt{learning rate:0.0002} &  \\
\texttt{loss function:BCE} & \texttt{optimizer:Adam} \\
\hline
\end{tabular}
\end{table}

\begin{figure}
\centering
\includegraphics[width=1.0\linewidth]{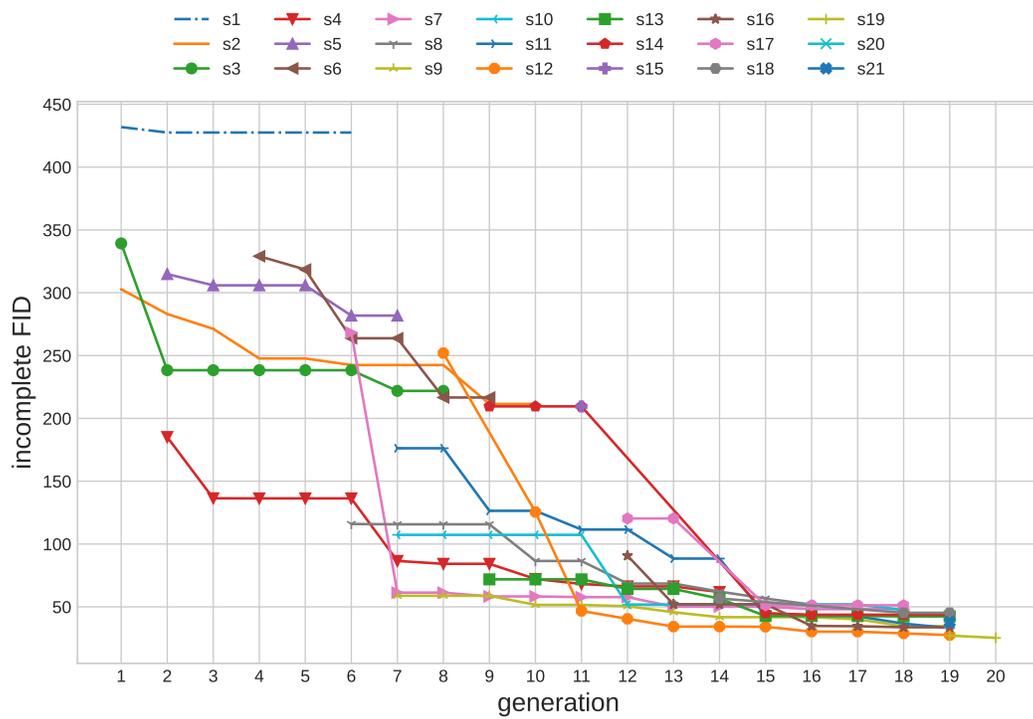}
\caption{The best incomplete FID trend of each species after being incompletely trained during SANE-GAN-M-e2. Each break line represents one species.}
\label{fig:dcganevo}
\end{figure}

We firstly conducted DCGAN-type architecture evolution experiment with MNIST for $3$ times, and generated $3$ edition architecture, i.e., SANE-GAN-M-e1, SANE-GAN-M-e2 and SANE-GAN-M-e3 as detailed in Table~\ref{tab:ganm}. Taking SANE-GAN-M-e2 as example, the evolution process went through $20$ generations, and the change process of the best incomplete FID in each species is shown in Figure~\ref{fig:dcganevo}. In early evolution process before generation $8$, the incomplete FIDs of species are relatively dispersed (range from $450$ to $50$). After generation $8$, the incomplete FIDs tend to convergence (less than $50$). It demonstrates that diversity protection is of great importance during evolution search. 

Both the scale and performance of DCGAN and GAN by SANE for MNIST and FashionMNIST are compared in Tables~\ref{tab:ganm} and \ref{tab:ganfm}, respectively. We recorded the total cell number (cell), total trainable parameter number (parameter) and the FID of GAN architectures. Besides, we recorded the evolution generation (gen) and time (time) of each experiment. The results show that with identical GAN components, DCGANs by SANE reach the similar FID compared to its competitor, while contain fewer trainable parameters. 

\begin{table}[!ht]
\caption{Scale and performance of GAN Architectures for MNIST\label{tab:ganm}}
\centering
\begin{tabular*}{1\textwidth}{@{\extracolsep{\fill}}ccccccc}
\toprule
{\scriptsize{}method} & {\scriptsize{}cell} & {\scriptsize{}layer} & {\scriptsize{}parameter} & {\scriptsize{}FID} & {\scriptsize{}gen} & {\scriptsize{}time}  \tabularnewline 
\midrule
{\scriptsize{}DCGAN \cite{radforc_unsupervised_2016}} & {\scriptsize{}4 convtrans + 4 conv} & {\scriptsize{}21} & {\scriptsize{}1.73M} & {\scriptsize{}16.95} & - & - \tabularnewline
{\scriptsize{}SANE-GAN-M-e1 (ours)} & {\scriptsize{}5 convtrans + 4 conv} & {\scriptsize{}25} & {\scriptsize{}\textbf{0.31M}} & {\scriptsize{}\textbf{16.71}} & {\scriptsize{}21} & {\scriptsize{}15.86h/GPU} \tabularnewline
{\scriptsize{}SANE-GAN-M-e2 (ours)} & {\scriptsize{}4 convtrans + 5 conv} & {\scriptsize{}25} & {\scriptsize{}\textbf{0.14M}} & {\scriptsize{}\textbf{15.86}} & {\scriptsize{}20} & {\scriptsize{}16.05h/GPU} \tabularnewline
{\scriptsize{}SANE-GAN-M-e3 (ours)} & {\scriptsize{}4 convtrans + 3 conv} & {\scriptsize{}19} & {\scriptsize{}\textbf{0.15M}} & {\scriptsize{}\textbf{16.21}} & {\scriptsize{}14} & {\scriptsize{}8.11h/GPU} \tabularnewline
\bottomrule
\end{tabular*}
\end{table}

\begin{table}[!ht]
\caption{Scale and performance of GAN for FashionMNIST\label{tab:ganfm}}
\centering
\begin{tabular*}{1\textwidth}{@{\extracolsep{\fill}}ccccccc}
\toprule
{\scriptsize{}method} & {\scriptsize{}cell} & {\scriptsize{}layer} & {\scriptsize{}parameter} & {\scriptsize{}FID} & {\scriptsize{}gen} & {\scriptsize{}time}  \tabularnewline 
\midrule
{\scriptsize{}DCGAN \cite{radforc_unsupervised_2016}} & {\scriptsize{}4 convtrans + 4 conv} & {\scriptsize{}21} & {\scriptsize{}1.73M} & {\scriptsize{}29.15} & - & - \tabularnewline
{\scriptsize{}SANE-GAN-F-e1 (ours)} & {\scriptsize{}5 convtrans + 5 conv} & {\scriptsize{}28} & {\scriptsize{}\textbf{0.52M}} & {\scriptsize{}\textbf{28.90}} & {\scriptsize{}18} & {\scriptsize{}21.05h/GPU} \tabularnewline
{\scriptsize{}SANE-GAN-F-e2 (ours)} & {\scriptsize{}5 convtrans + 4 conv} & {\scriptsize{}25} & {\scriptsize{}\textbf{0.75M}} & {\scriptsize{}\textbf{28.19}} & {\scriptsize{}20} & {\scriptsize{}28.94h/GPU} \tabularnewline
{\scriptsize{}SANE-GAN-F-e3 (ours)} & {\scriptsize{} 5 convtrans +  5 conv} & {\scriptsize{}28} & {\scriptsize{}\textbf{0.53M}} & {\scriptsize{}\textbf{28.72}} & {\scriptsize{}22} & {\scriptsize{}30.26h/GPU} \tabularnewline
\bottomrule
\end{tabular*}
\end{table}

\begin{figure*}[htbp]
\centering
\includegraphics[width=0.43\linewidth]{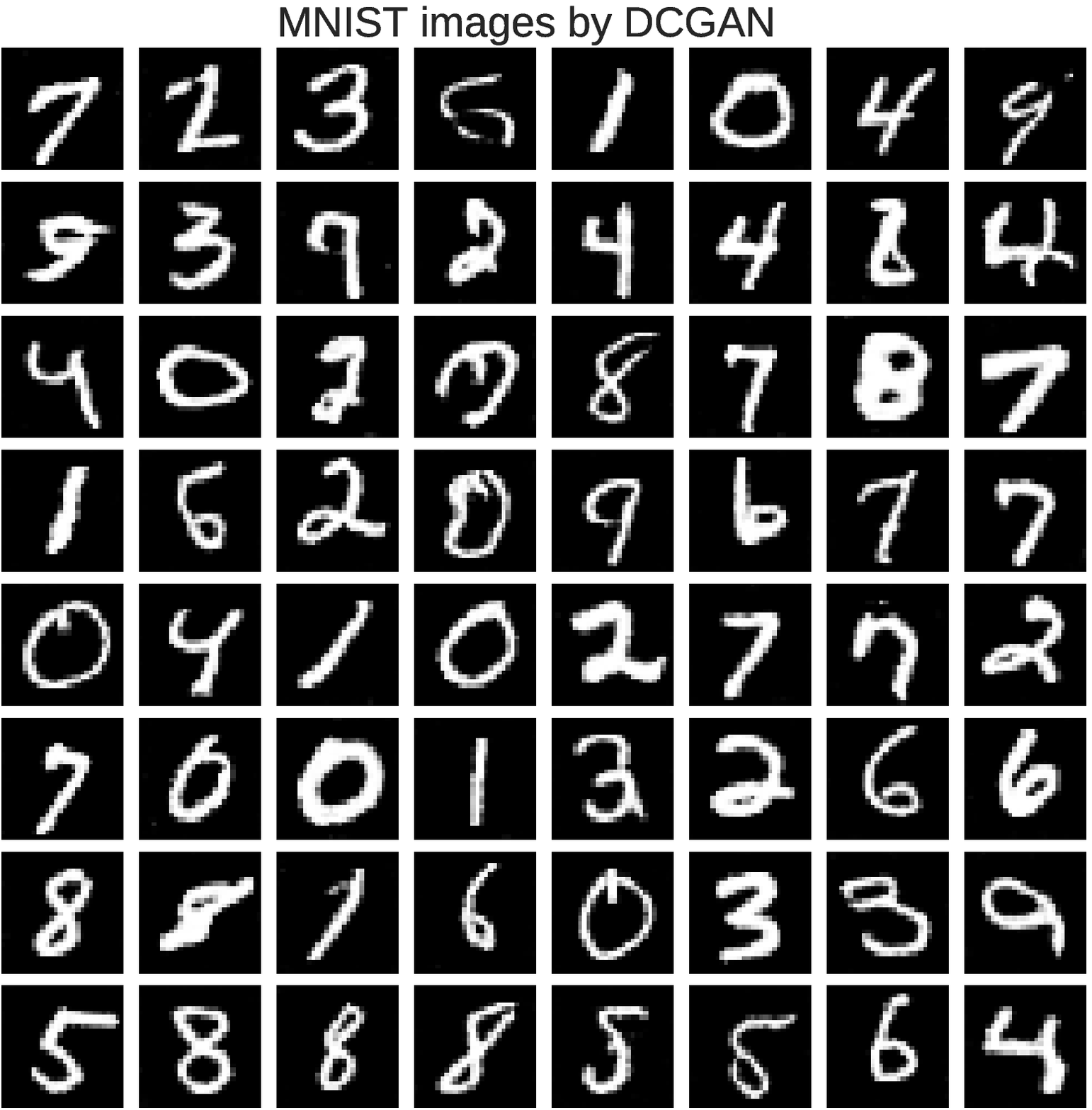}
\includegraphics[width=0.43\linewidth]{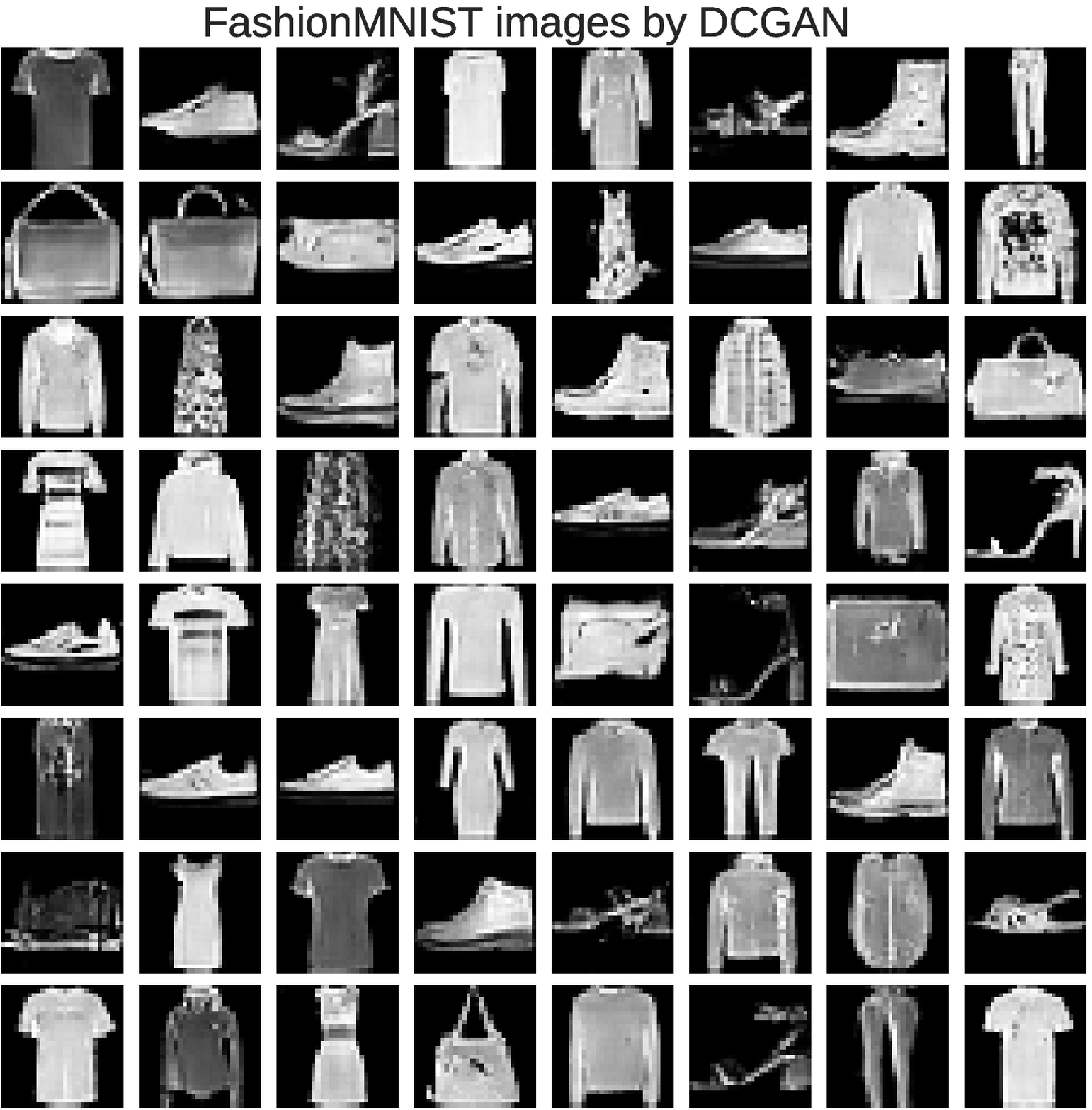}
\includegraphics[width=0.43\linewidth]{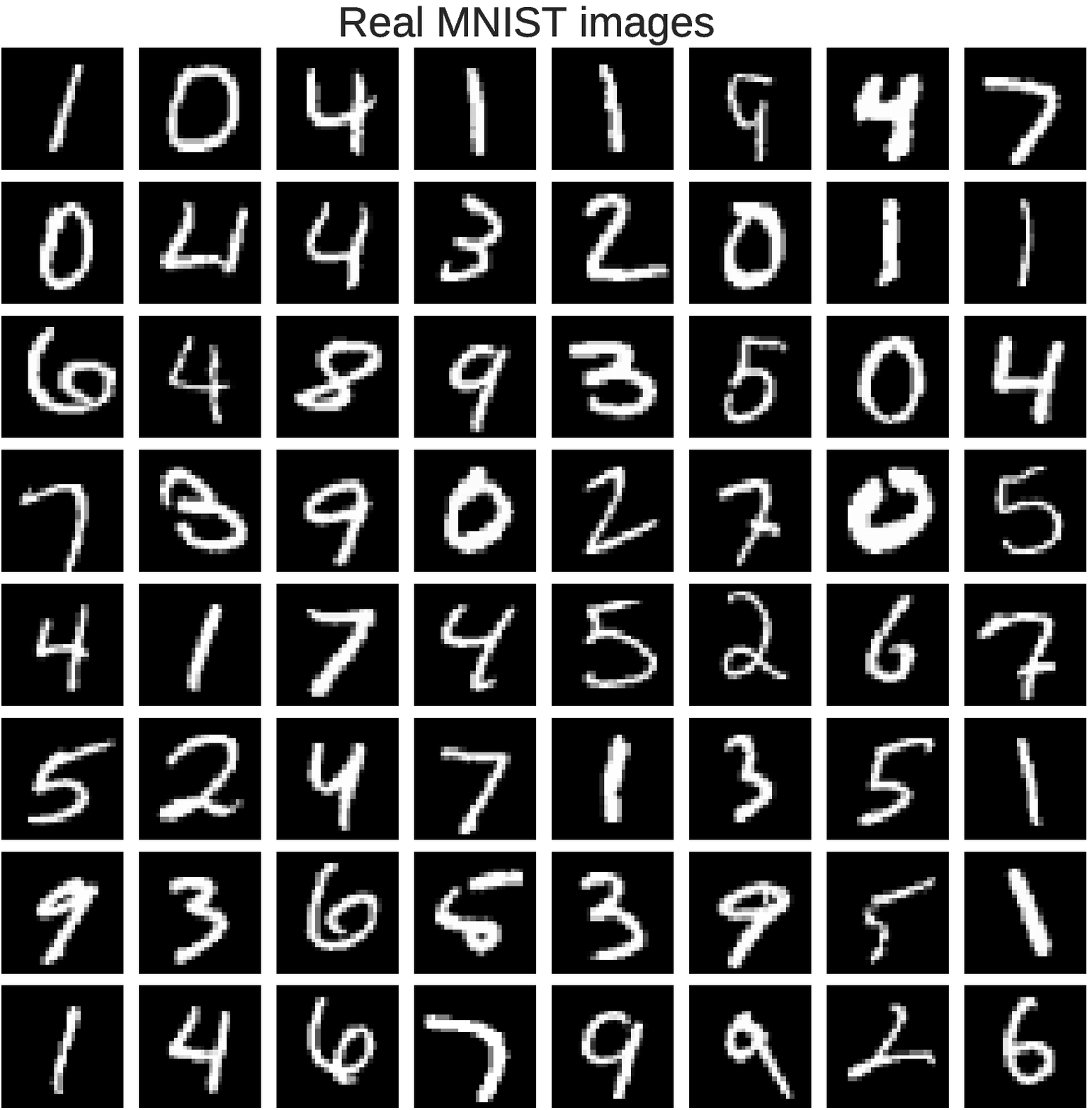}
\includegraphics[width=0.43\linewidth]{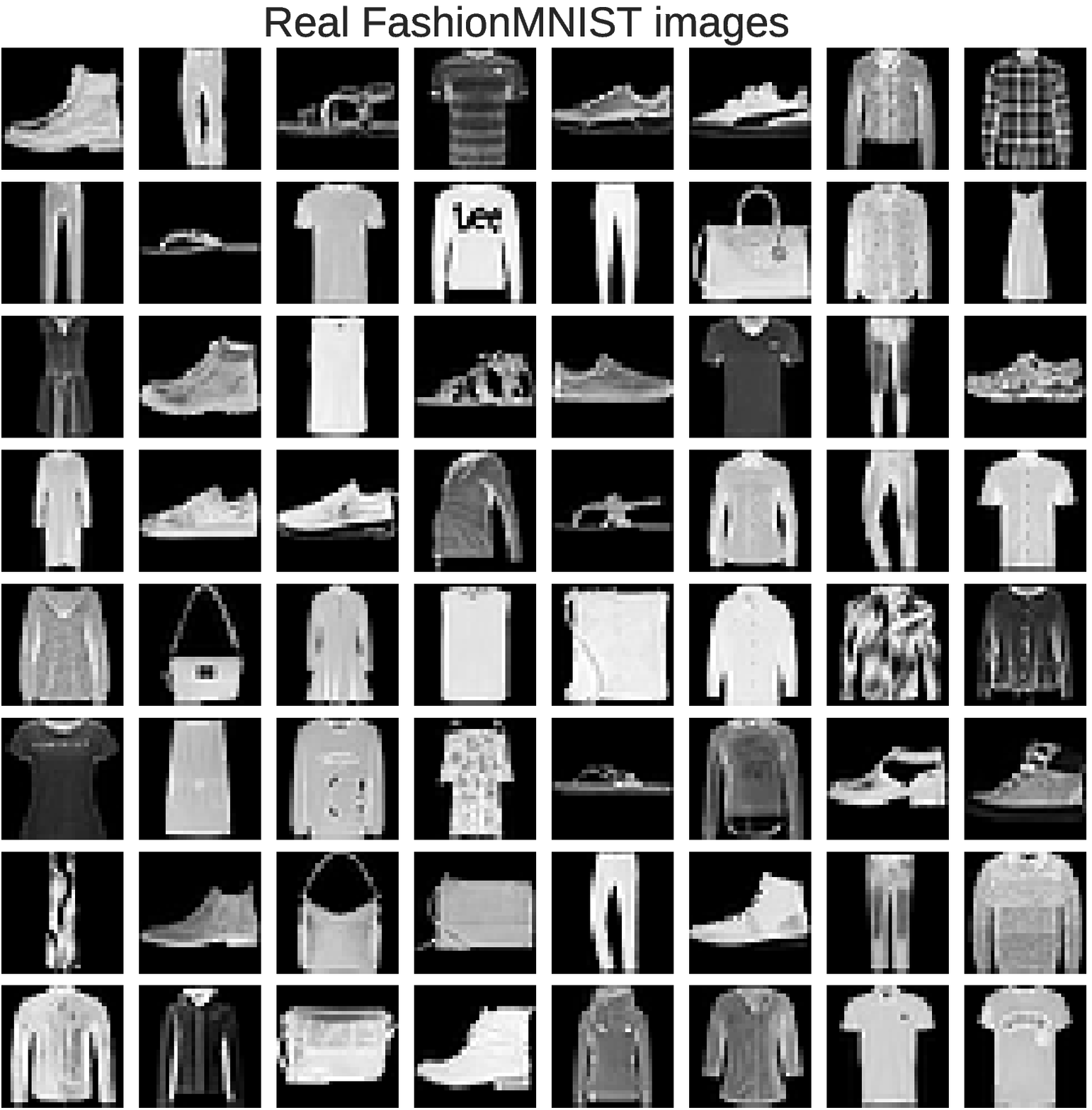}
\includegraphics[width=0.43\linewidth]{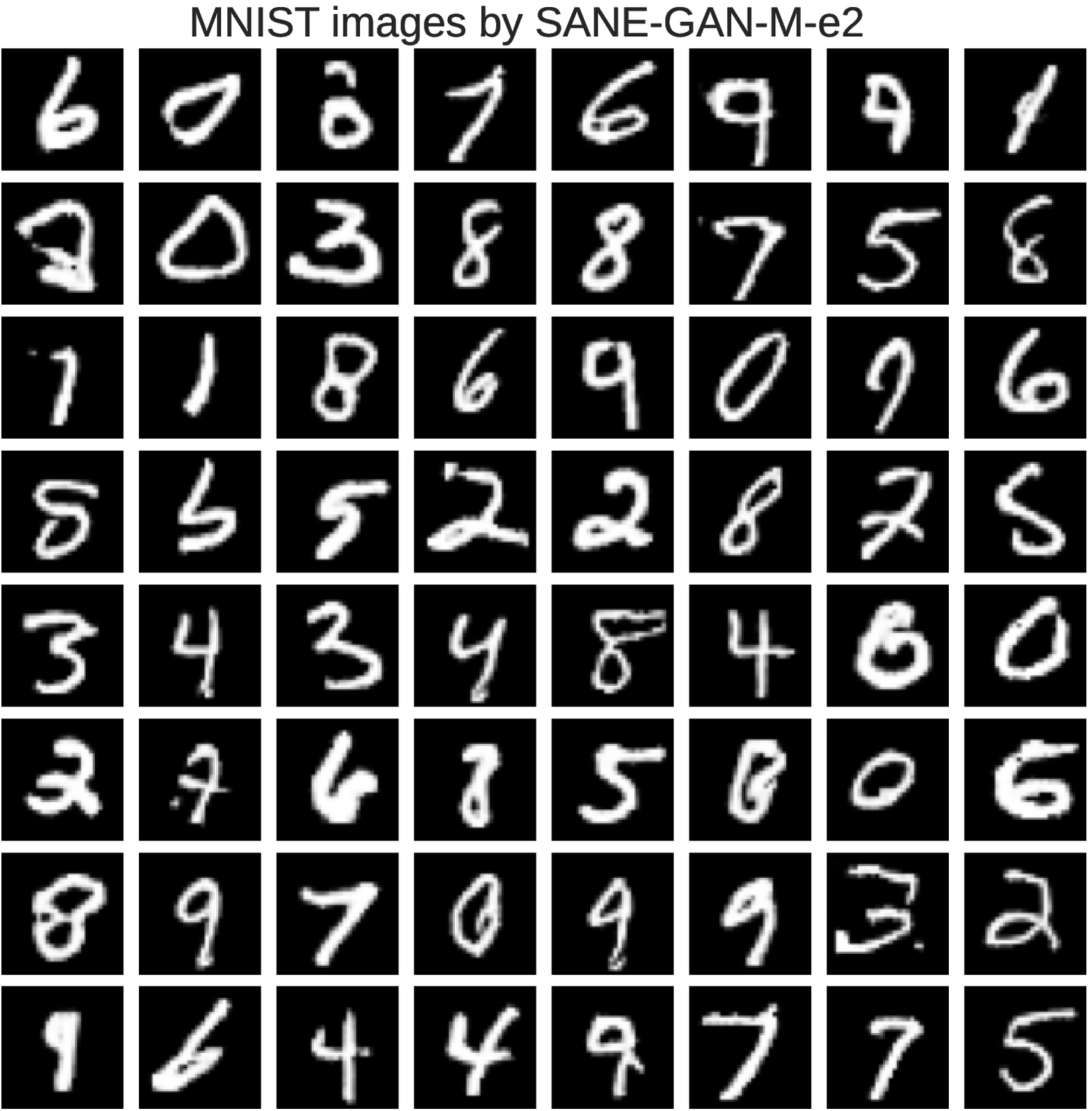}
\includegraphics[width=0.43\linewidth]{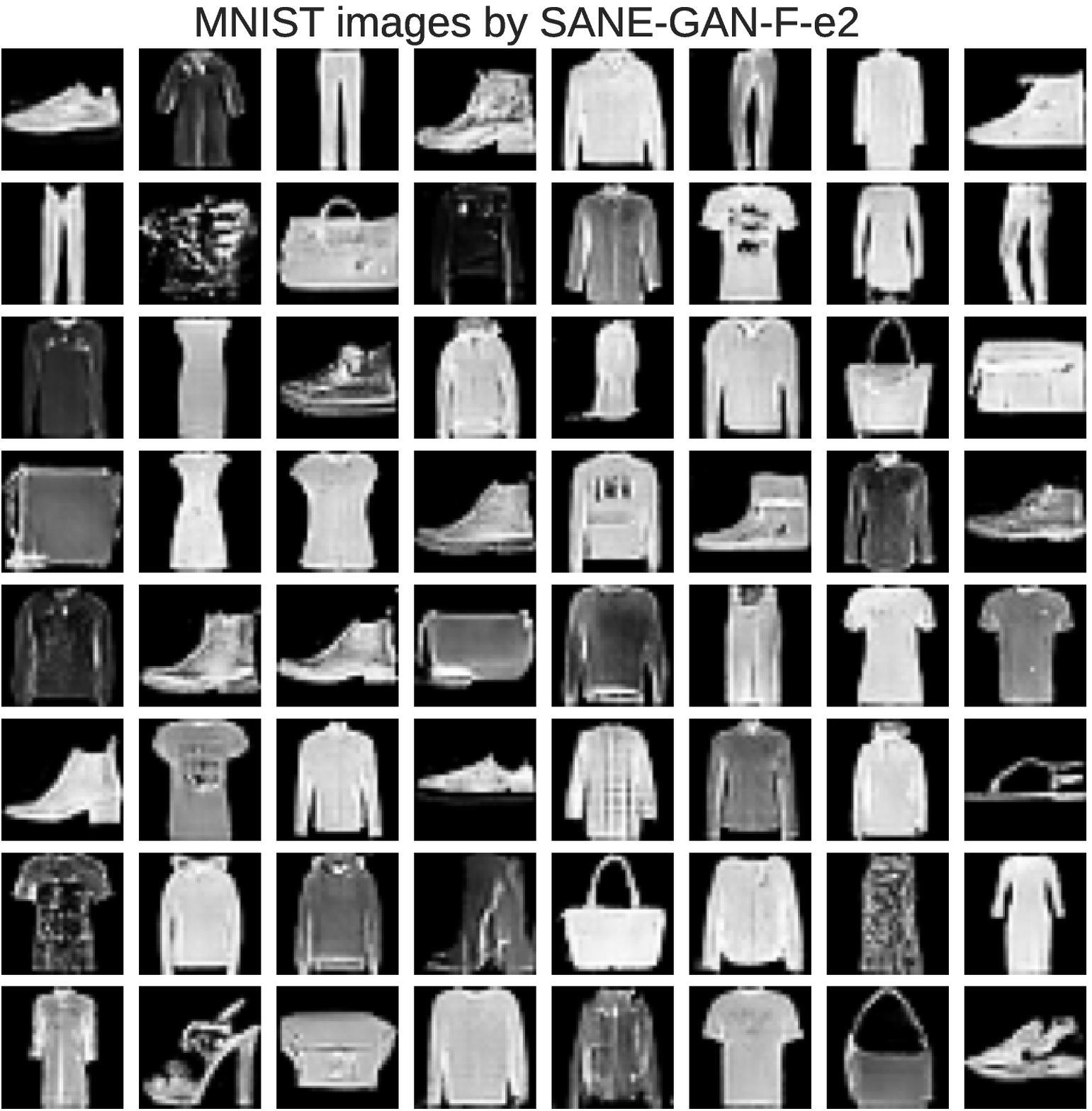}
\caption{Real vs generated image samples. From top to bottom in two columns, they are digit and clothes images from DCGAN, dataset and SANE-GAN-M-e2/SANE-GAN-F-e2}
\label{fig:realandfake}
\end{figure*}

Figure~\ref{fig:realandfake} displays some images from dataset, as well as some images generated by DCGAN, SANE-GAN-M-e2 and SANE-GAN-F-e2. In MNIST column (left), we can see that both DCGAN and SANE-DCGAN-M-e2 are able to capture the features of digit images in MNIST dataset and generate realistic digit images except for a few unrecognizable symbol image. In FashionMNIST column (right), DCGAN and SANE-GAN-F-e2 perform equally well, and they are both capable of generating diverse clothes images including some with unusual designs.

\subsection{LSTM Evolution}

In LSTM evolution, we estimated LSTM performance by predicting image sequences in MovingMNIST dataset. MovingMNIST contains $8,000$ training image sequences and $2,000$ test image sequences, each of which is $20$ frames long and contains $2$ handwritten digits bouncing inside a $64\times 64$ patch. The moving digits in the image sequences exhibits complicated appearance, such as occlusion and bounce during the movement. It is hard for LSTM to give accurate predictions on test set without learning the inner dynamics of the MovingMNIST. The performance metric of LSTM architectures during evolution is BCE loss for predicting the last $10$ frames for each sequence in MovingMNIST conditioned on the first $10$ frames.

In our experiment, LSTM for sequence prediction contains one encoder organ which is used to encode a sequence into a fixed length representation, and one decoder organ which is used to decode a sequence out of that representation. LSTM cell in encoder output hidden state and cell state as input to corresponding LSTM cell in decoder, so encoder architecture determines decoder architecture in a large degree. We evolve encoder organ in LSTM with only mutation operations, since there exist strong coupling relation between encoder and decoder organ.

We chose ConvLSTM \cite{shi_convlstm_2015} which is suitable for image sequence prediction as the LSTM competitor. It is composed by convolution cell, convlution transpose cell and LSTM cell, so we set the type-free search space as Eq.~\eqref{eq:slstm}, where $e_{m}$ represent LSTM cell, organ $o_{e}$ and $o_{r}$ represent encoder and decoder organ respectively.

\begin{equation}
\mathbb{S}_{LSTM}=
\begin{cases}
    \boldsymbol{E}=\{e_{c},e_{t},e_{m}\}, & \boldsymbol{E}^{(o)}\subseteq\boldsymbol{E};\\
    \boldsymbol{O}=\{o_{e},o_{r}\}, & \boldsymbol{E}^{(o_{e})}=\{e_{c},e_{m}\},\boldsymbol{E}^{(o_{r})}=\{e_{t},e_{m}\};\\
    \boldsymbol{R}=\{r_{d},r_{r}\}, & r_{d}=1,r_{r}=\{(o_{e}, o_{r});\\
    &(e_{c},e_{c}), (e_{t}, e_{t}), (e_{m}, e_{m}), (e_{c}, e_{m}),\\
    &(e_{m}, e_{c}), (e_{m},e_{t}), (e_{t}, e_{m})\}.
\end{cases}\label{eq:slstm}
\end{equation}

Table~\ref{tab:lstmevo} generalizes key evolution configuration in regard to LSTM attributions, training and evolution. Most configurations are the same with these in CNN evolution. The convolution LSTM cells (ConvLSTM) in same stage from encoder and decoder organ are identical, they contain a convolution core module and a group normalization affiliated module.

\begin{table}[!ht]
\caption{LSTM Evolution Configurations}\label{tab:lstmevo}
\centering
\begin{tabular}{ |ll| } 
\hline
\multicolumn{2}{|c|}{DNN config} \\ 
\hline
\texttt{DNN type:LSTM} & \texttt{organ types:[encoder]} \\ 
\texttt{encoder cell types:} &  \\
\multicolumn{2}{|l|}{\texttt{\{conv:[[16,3,1,0],leakyrelu],}} \\
\multicolumn{2}{|l|}{\texttt{convlstm:[3,groupnorm]\}}} \\
\hline
\hline
\multicolumn{2}{|c|}{evolution config} \\
\hline
\texttt{organ prob:[100\%]} & \\
\multicolumn{2}{|l|}{\texttt{conv attr prob:[40\%,30\%,20\%,10\%]}} \\
\multicolumn{2}{|l|}{\texttt{conv attr growth factor:[16,2,1,1]}} \\
\multicolumn{2}{|l|}{\texttt{convlstm attr growth factor:2}} \\
\multicolumn{2}{|l|}{\texttt{species num limit:10}} \\
\multicolumn{2}{|l|}{\texttt{species distance threshold:1.0}} \\
\hline
\hline
\multicolumn{2}{|c|}{training config} \\
\hline
\texttt{train rate:50\%} &  \\
\texttt{incomplete train epochs:5} & \texttt{complete train epochs:20} \\
\texttt{train batches:25} & \texttt{learning rate:0.0001} \\
\texttt{loss function:BCE} & \texttt{optimizer:Adam} \\
\hline
\end{tabular}
\end{table}

Both the scale and performance of ConvLSTM and LSTM by SANE are compared in Table~\ref{tab:convlstm}. We recorded total cell number (cell), total trainable parameter number (parameter) and the BCE loss of LSTMs. Besides, we recorded the evolution generation (gen) and time (time) of each experiment. The results show that, with identical LSTM components, ConvLSTMs by SANE reach the similar BCE loss compared to its competitor, while contain fewer cells and trainable parameters. 

\begin{table}[!ht]
\caption{Scale and performance of LSTMs for MovingMNIST
\label{tab:convlstm}}

\centering{}%
\begin{tabular*}{1\textwidth}{@{\extracolsep{\fill}}ccccccc}
\toprule
{\scriptsize{}method} & {\scriptsize{}cell} & {\scriptsize{}layer} & {\scriptsize{}parameter} & {\scriptsize{}BCE loss} & {\scriptsize{}gen} & {\scriptsize{}time}  \tabularnewline 

\midrule
 & {\scriptsize{}5 conv +} & & & & & \tabularnewline
{\scriptsize{}ConvLSTM \cite{shi_convlstm_2015}} & {\scriptsize{}2 convtras + } & {\scriptsize{}56} & {\scriptsize{}9.02M} & {\scriptsize{}0.0705} & - & - \tabularnewline
 & {\scriptsize{}6 convlstm} & & & & & \tabularnewline
 
 & {\scriptsize{}1 conv +} & & & & & \tabularnewline
{\scriptsize{}SANE-ConvLSTM-e1 (ours)} & {\scriptsize{}1 convtras + } & {\scriptsize{}46} & {\scriptsize{}\textbf{6.58M}} & {\scriptsize{}\textbf{0.0700}} & {\scriptsize{}13} & {\scriptsize{}32.72h/GPU} \tabularnewline
 & {\scriptsize{}6 convlstm} & & & & & \tabularnewline
 
 & {\scriptsize{}1 conv +} & & & & & \tabularnewline
{\scriptsize{}SANE-ConvLSTM-e2 (ours)} & {\scriptsize{}1 convtras + } & {\scriptsize{}32} & {\scriptsize{}\textbf{8.52M}} & {\scriptsize{}\textbf{0.0698}} & {\scriptsize{}11} & {\scriptsize{}37.85h/GPU} \tabularnewline
 & {\scriptsize{}4 convlstm} & & & & & \tabularnewline 
 
 & {\scriptsize{}3 conv +} & & & & & \tabularnewline
{\scriptsize{}SANE-ConvLSTM-e3 (ours)} & {\scriptsize{}3 convtras + } & {\scriptsize{}34} & {\scriptsize{}\textbf{4.08M}} & {\scriptsize{}\textbf{0.0712}} & {\scriptsize{}15} & {\scriptsize{}35.33h/GPU} \tabularnewline
 & {\scriptsize{}4 convlstm} & & & & & \tabularnewline 
 
\bottomrule
\end{tabular*}
\end{table}

\begin{figure*}[htbp]
	\centering
	\includegraphics[width=0.8\textwidth]{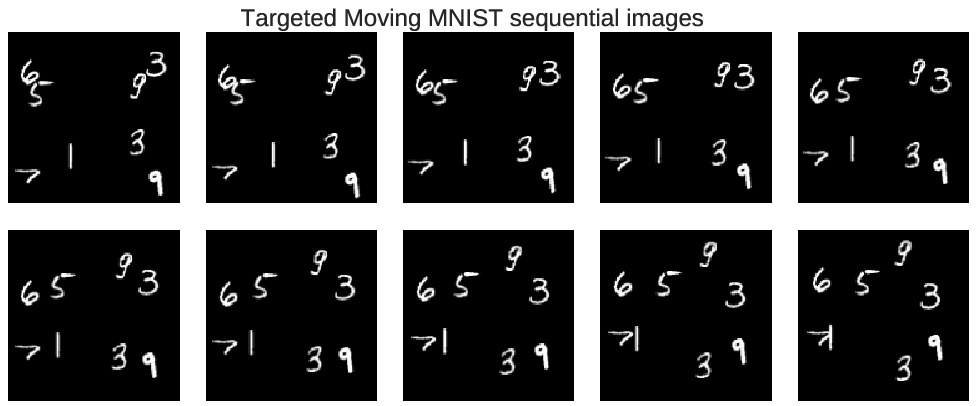}
    \includegraphics[width=0.8\textwidth]{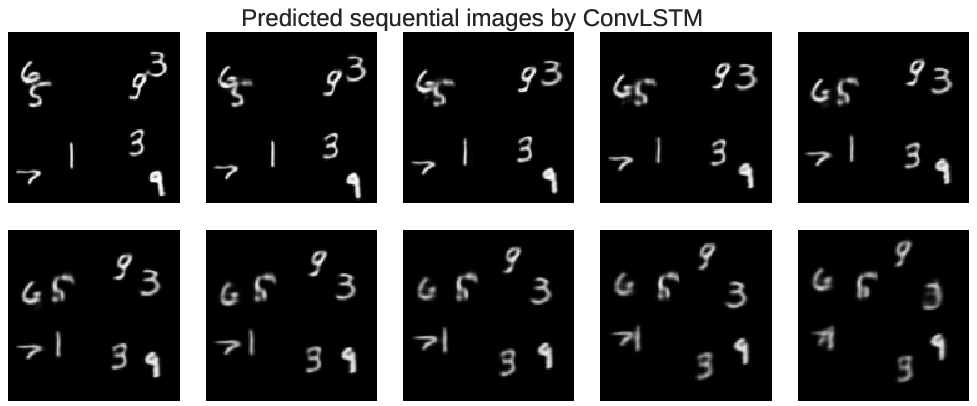}
    \includegraphics[width=0.8\textwidth]{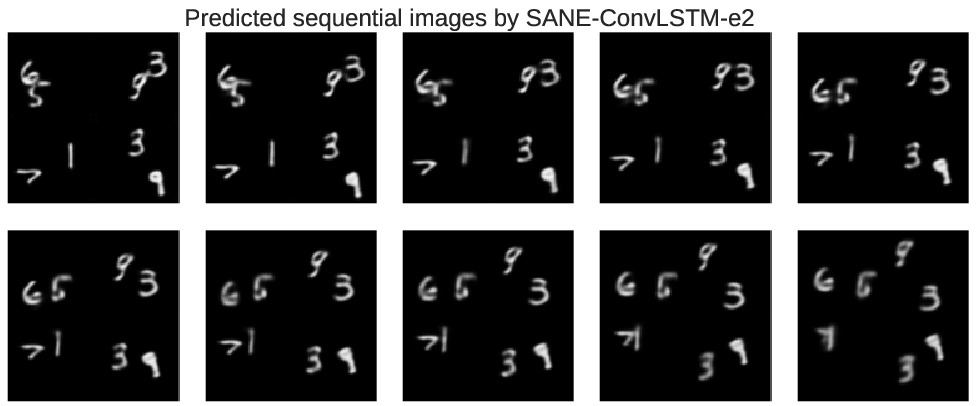}

	\caption{Targeted MovingMNIST sequential images and predicted sequential images accordingly by LSTMs. Randomly choose 4 groups of sequential images from test dataset in MovingMNIST, input first 10 images in each group to LSTMs to predict the following 10 sequential images, and use last 10 ones in each group as targets.}
\label{fig:clstmcontrast}
\end{figure*}

Figure~\ref{fig:clstmcontrast} displays some images from dataset as well as some images generated by ConvLSTM and SANE-ConvLSTM-e2. From top to bottom, the three group shows target image sequence and predicted image sequence by ConvLSTM and SANE-ConvLSTM-e2. Both of the two LSTM can accurately predict the digits and their position in each frame in sequence.

\subsection{Comparison with State-of-the-art}

Most NAS studies are restrained within single DNN type as shown in Table~\ref{tab:state}. Researchers have taken much effort to improve prediction performance for specific DNN types and made many great achievements. Especially in CNN, CNN-GA \cite{sun_cnnga_2020} takes $37.5$day/GPU to construct a CNN with $2.9$M parameters that reaches $95.2$\% accuracy on CIFAR10. SI-EvoNet-S \cite{zhang_escnn_2021} imports attention mechanism into CNN and takes $0.46$day/GPU to obtain a CNN with $1.84$M parameters that reaches $97.31$\% accuracy on CIFAR10. These task-oriented methods are not easy to seamlessly transfer across different DNN types. Furthermore, blind pursuit of performance without considering the scale of the obtained DNN architectures is not beneficial to the balance between its performance and scale in practice. For AutoGAN \cite{gong_autogan_2019} and AdversarialNAS \cite{gao_anas_2020}, their experimental results show great performance on CIFAR10, but the scale of the obtained GAN architecture is not recorded. There are relatively few NAS studies for LSTM, ENAS-LSTM \cite{hamdi_elstm_2021} constructs LSTM for text generation task and compares with only hand-designed vanilla LSTM. Perplexity (ppl), a commonly used metric for evaluating the efficacy of generative models, is used as a measure of probability for a sentence to be produced by the model trained on a dataset. 

Compared with the existing methods, SANE is able to construct $3$ types of DNN type, i.e., CNN, GAN and LSTM, by uniform search space and search strategy. Besides, as a constructive neuroevolution method, SANE grows DNN architectures only when it improves their fitness. This evolution scheme keeps the scale of the obtained DNN architectures near-minimal during evolution search while saving enormous computation resources. All SANE experiments take less than $2.5$day/GPU with $1$ RTX 3090 GPU ($24$GB).

\begin{table}[!ht]
\caption{SANE and State-of-the-art methods}\label{tab:state}
\centering
\begin{tabular*}{1\textwidth}{@{\extracolsep{\fill}}ccccccc}
\toprule
{\scriptsize{}method} & {\scriptsize{}DNN type} & {\scriptsize{}parameter} & {\scriptsize{}dataset} & {\scriptsize{}performance} & {\scriptsize{}time} \tabularnewline 
\midrule
{\scriptsize{}} & {\scriptsize{}CNN} & {\scriptsize{}1.35M} & {\scriptsize{}CIFAR10} & {\scriptsize{}acc 91.80\%} & {\scriptsize{}0.81day/GPU} \tabularnewline
{\scriptsize{}SANE (ours)} & {\scriptsize{}GAN} & {\scriptsize{}0.75M} & {\scriptsize{}MNIST} & {\scriptsize{}FID 15.86} & {\scriptsize{}0.67day/GPU} \tabularnewline
{\scriptsize{}} & {\scriptsize{}LSTM} & {\scriptsize{}6.58M} & {\scriptsize{}MovingMNIST} & {\scriptsize{}BCE loss 0.0700} & {\scriptsize{}1.36day/GPU} \tabularnewline
\midrule
{\scriptsize{}CNN-GA \cite{sun_cnnga_2020}} & {\scriptsize{}CNN} & {\scriptsize{}2.9M} & {\scriptsize{}CIFAR10} & {\scriptsize{}acc 95.2\%} & {\scriptsize{}37.5day/GPU} \tabularnewline
\midrule
{\scriptsize{}SI-EvoNet-S \cite{zhang_escnn_2021}} & {\scriptsize{}CNN} & {\scriptsize{}1.84M} & {\scriptsize{}CIFAR10} & {\scriptsize{}acc 97.31\%} & {\scriptsize{}0.46day/GPU} \tabularnewline
\midrule
{\scriptsize{}AutoGAN \cite{gong_autogan_2019}} & {\scriptsize{}GAN} & {\scriptsize{}-} & {\scriptsize{}CIFAR10} & {\scriptsize{}FID 12.42} & {\scriptsize{}2day/GPU} \tabularnewline
\midrule
{\scriptsize{}AdversarialNAS \cite{gao_anas_2020}} & {\scriptsize{}GAN} & {\scriptsize{}-} & {\scriptsize{}CIFAR10} & {\scriptsize{}FID 10.87} & {\scriptsize{}1day/GPU} \tabularnewline
\midrule
{\scriptsize{}ENAS-LSTM \cite{hamdi_elstm_2021}} & {\scriptsize{}LSTM} & {\scriptsize{}-} & {\scriptsize{}Penn TreeBank} & {\scriptsize{}ppl 319} & {\scriptsize{}1day/GPU} \tabularnewline
\bottomrule
\end{tabular*}
\end{table}

\section{Conclusion \label{sub:conclusion}}

In this study, we proposed a self-adaptive neuroevolution method (SANE) to automatically construct DNN architectures. To investigate SANE, we conducted CNN, GAN and LSTM architectures evolution search experiments. The search process illustrated that the type-free search space was capable of applying to different DNN types, and the self-adaptive adjustment mechanism was effective to regulate evolution search setting to adapt to changing search state. The search results verified that constructive evolution strategy in SANE was able to design DNN architectures of different types with near-minimal scale for given performance standard.

SANE has been successfully applied to construct existing DNNs with conventional composition and integration, such as chain-structured CNN. However, SANE sets the type-free search space based on existing DNN types, we didn't consider modifying their overall frameworks and basic components. In future work, the type-free search space in SANE could be extend by importing cells with novel structure and connection patterns in geometry form to explore new DNN types. More universal evolution operations in a large range of granularity would be designed to improve evolution efficiency or population diversity.

\section{Acknowledgment}
The authors sincerely thank Prof. Danilo R. B. Araujo, the editors and the anonymous reviewers for their very helpful suggestions that have improved the presentation of our paper.
This work is supported by the National Natural Science Foundation of China (Grant Nos. 62176036, 62102059), the Liaoning Collaborative Fund (Grant No. 2020-HYLH-17), and the Fundamental Research Funds for the Central Universities (No. 3132022225).

\bibliographystyle{elsarticle-num}
\bibliography{SANE}



\end{document}